\documentclass[sigconf]{aamas}

\graphicspath{{figures/}}
\usepackage{balance} 
\usepackage{algorithmic}
\usepackage{algorithm}
\usepackage{multirow}
\usepackage{makecell}

\makeatletter
\gdef\@copyrightpermission{
  \begin{minipage}{0.2\columnwidth}
   \href{https://creativecommons.org/licenses/by/4.0/}{\includegraphics[width=0.90\textwidth]{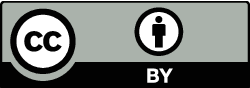}}
  \end{minipage}\hfill
  \begin{minipage}{0.8\columnwidth}
   \href{https://creativecommons.org/licenses/by/4.0/}{This work is licensed under a Creative Commons Attribution International 4.0 License.}
  \end{minipage}
  \vspace{5pt}
}
\makeatother

\setcopyright{ifaamas}
\acmConference[AAMAS '25]{Proc.\@ of the 24th International Conference
on Autonomous Agents and Multiagent Systems (AAMAS 2025)}{May 19 -- 23, 2025}
{Detroit, Michigan, USA}{Y.~Vorobeychik, S.~Das, A.~Nowé  (eds.)}
\copyrightyear{2025}
\acmYear{2025}
\acmDOI{}
\acmPrice{}
\acmISBN{}

\acmSubmissionID{<<OpenReview submission id>>}

\title[AAMAS-2025 Formatting Instructions]{Salience-Invariant Consistent Policy Learning for Generalization in Visual Reinforcement Learning}

\author{Jingbo Sun}
\affiliation{
  \institution{Institute of Automation, CASIA}
  \city{Beijing}
  \country{China} \\
  \institution{Pengcheng Laboratory}
  \city{Shenzhen}
  \country{China} \\
  \institution{School of Artificial Intelligence, UCAS}
  \city{Beijing}
  \country{China}}
\email{sunjingbo2022@ia.ac.cn}

\author{Songjun Tu}
\affiliation{
  \institution{Institute of Automation, CASIA}
  \city{Beijing}
  \country{China} \\
  \institution{Pengcheng Laboratory}
  \city{Shenzhen}
  \country{China} \\
  \institution{School of Artificial Intelligence, UCAS}
  \city{Beijing}
  \country{China}}
\email{tusongjun2023@ia.ac.cn}

\author{Qichao Zhang}
\affiliation{
  \institution{Institute of Automation, CASIA}
  \city{Beijing}
  \country{China} \\
  \institution{School of Artificial Intelligence, UCAS}
  \city{Beijing}
  \country{China}}
\email{zhangqichao2014@ia.ac.cn}
\thanks{Corresponding author: Qichao Zhang, Dongbin Zhao.}

\author{Ke Chen}
\affiliation{
  \institution{Pengcheng Laboratory}
  \city{Shenzhen}
  \country{China}}
\email{chenk02@pcl.ac.cn}

\author{Dongbin Zhao}
\affiliation{
  \institution{Institute of Automation, CASIA}
  \city{Beijing}
  \country{China} \\
  \institution{School of Artificial Intelligence, UCAS}
  \city{Beijing}
  \country{China}}
\email{dongbin.zhao@ia.ac.cn}

\begin{abstract}
Generalizing policies to unseen scenarios remains a critical challenge in visual reinforcement learning, where agents often overfit to the specific visual observations of the training environment. 
In unseen environments, distracting pixels may lead agents to extract representations containing task-irrelevant information.  
As a result, agents may deviate from the optimal behaviors learned during training, thereby hindering visual generalization.
To address this issue, we propose the \textbf{S}alience-Invariant \textbf{C}onsistent \textbf{P}olicy \textbf{L}earning (SCPL) algorithm, an efficient framework for zero-shot generalization. 
Our approach introduces a novel value consistency module alongside a dynamics module to effectively capture task-relevant representations.
The value consistency module, guided by saliency, ensures the agent focuses on task-relevant pixels in both original and perturbed observations, while the dynamics module uses augmented data to help the encoder capture dynamic- and reward-relevant representations.
Additionally, our theoretical analysis highlights the importance of policy consistency for generalization. 
To strengthen this, we introduce a policy consistency module with a KL divergence constraint to maintain consistent policies across original and perturbed observations.
Extensive experiments on the DMC-GB, Robotic Manipulation, and CARLA benchmarks demonstrate that SCPL significantly outperforms state-of-the-art methods in terms of generalization. Notably, SCPL achieves average performance improvements of 14\%, 39\%, and 69\% in the challenging DMC video hard setting, the Robotic hard setting, and the CARLA benchmark, respectively.
Project Page: \url{https://sites.google.com/view/scpl-rl}.
\end{abstract}


\newcommand{\BibTeX}{\rm B\kern-.05em{\sc i\kern-.025em b}\kern-.08em\TeX}

\begin{document}


\pagestyle{fancy}
\fancyhead{}


\maketitle 


\section{Introduction}
In recent years, visual reinforcement learning (RL) \cite{sutton_rl} has achieved remarkable success across various domains, including video games \cite{game1, tang2020}, robot control \cite{li2019deep, cp3er, tu2024dataset}, and autonomous driving \cite{drive2, wang2021highway, sun2023reinforcement}.
However, generalizing policies to novel scenarios remains a significant challenge. 
Small visual perturbations in observations can distract RL agents \cite{liu2022soft,madi}, leading to representations containing task-irrelevant information and decisions that deviate from their training behavior, ultimately hindering visual generalization.
In this paper, we aim to develop generalizable RL agents that can generate effective task-relevant representations and make consistent decisions across both original and perturbed observations.

\begin{figure*}
   \centering
   \includegraphics[width=\linewidth]{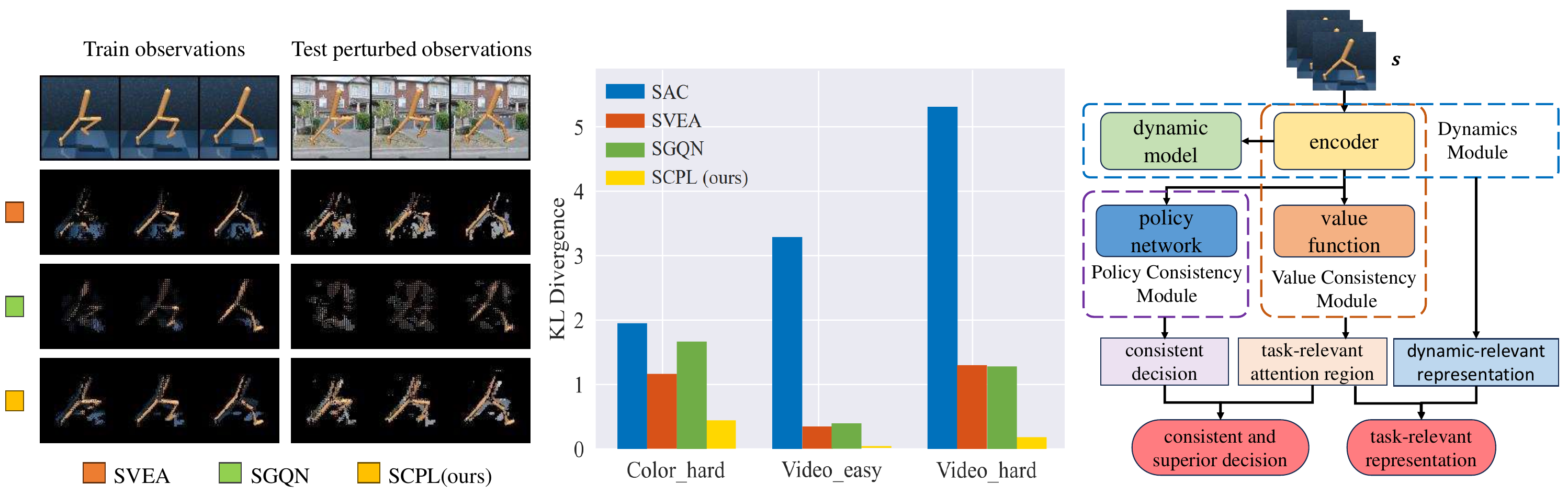}
  \caption{(Left) Saliency masked map of SVEA, SGQN, and SCPL (ours), which shows the attention regions of value functions on the DMC-GB benchmark. 
  (Middle) The KL divergence of action distribution between training and test environments on DMC-GB, where our method holds the smallest KL divergence. 
  (Right) Contribution overview of SCPL, which aims to improve visual generalization by achieving task-relevant representations and consistent and superior decisions.
  } 
  \label{fig_first}
  \Description{}
  \vspace{-.2cm}
\end{figure*}

Data augmentation (DA)-based methods \cite{rad,drq,srm} are widely used to enhance the representational ability of visual RL agents.
Recent advances, such as SVEA \cite{svea} and SGQN \cite{sgqn}, leverage augmented data for implicit regularization to improve generalization. 
Unfortunately, as illustrated in Fig.\ref{fig_first} (left),these methods fail to maintain consistent task-relevant attention regions in perturbed observations, impeding the learning of task-relevant representations.
Other studies \cite{dbc, spr} employ dynamic models as auxiliary tasks to capture task-relevant representations.
However, the encoder's primary design for original observations prevents it from generating task-relevant representations for perturbed observations.
Furthermore, it is difficult to generate task-relevant representations with uncritical attention to task-relevant regions.

Additionally, our theoretical analysis reveals that policy consistency between environments with original and perturbed observations is crucial for generalization.
However, previous methods focus on improving representations, often neglecting policy consistency.
As shown in Fig.\ref{fig_first} (middle), both SVEA \cite{svea} and SGQN \cite{sgqn} exhibit high KL divergence in action distributions between training and test environments, indicating inconsistent decisions across original and perturbed observations.
Improving the consistency of action distributions between original and perturbed observations is also essential to improve generalization.
These insights prompt us to consider: 
\textbf{Can we develop generalizable agents that maintain consistent task-relevant representations and policies?}

To address these challenges, we propose Salience-Invariant Consistent Policy Learning (\textbf{SCPL}), which improves generalization by encouraging RL agents to capture consistent task-relevant representations and make consistent superior decisions across diverse observations. 
First, we introduce a novel value consistency module that encourages the encoder and value function to capture \textbf{task-relevant attention region} in  observations.
Meanwhile, we introduce a dynamics module to generate \textbf{dynamic-relevant representations} for observations.
By combining the value consistency module and dynamics module, SCPL produces consistent task-relevant representations.
Furthermore, SCPL regularizes the policy network using a KL divergence constraint between the policies for original and augmented observations, enabling agents to make \textbf{consistent decisions} in test environments.
An overview of the motivation behind SCPL is illustrated in Fig.\ref{fig_first} (right).

In summary, our contribution includes the following aspects: 
\vspace{-.2cm}
\begin{itemize}
\item We propose an effective generalization framework, SCPL, in which a novel value consistency module with saliency guidance and a dynamics module with augmented data is introduced to generate task-relevant representations.
\item Through theoretical analysis, we reveal that improved policy consistency leads to enhanced generalization.
We further introduce a policy consistency module to regularize the policy network for consistent decisions to improve generalization. 

\item  The proposed SCPL achieves state-of-the-art (SOTA) performances in 3 popular visual generalization benchmarks, with an average boost of 14\%, 39\%, and 69\% on the \textit{video hard} setting in DMC-GB, the Robotic hard setting, and the CARLA benchmark, respectively.
\end{itemize}


\section{Related Works} \label{sec2}
\subsection{Data augmentation for visual RL} 
Data augmentation is widely used to enhance the generalization of visual reinforcement learning (RL) \cite{augmentation1,drg,rad}. 
DrQ \cite{drq} employs image transformation strategies to augment observations through implicit regularization. 
SVEA \cite{svea} enhances generalization by updating the value function with both original and augmented data. 
CG2A \cite{cg2a} improves generalization by combining various data augmentations and alleviating the gradient conflict bias caused by these augmentations. 
CNSN \cite{cnsn} employs normalization techniques to improve visual generalization. 
SGQN \cite{sgqn} utilizes saliency guidance to focus agents' attention on task-relevant areas in original observations while aligning their attention across original and augmented data using a trainable network.
MaDi \cite{madi} improves generalization by incorporating a mask network before the value function to filter out task-irrelevant regions in observations.
While these methods can identify effective task-relevant regions in original observations, they often struggle with perturbed observations. In this paper, SCPL focuses on maintaining consistent task-relevant regions in both original and perturbed data with a value consistency module.

\begin{figure*}
  \centering
\includegraphics[width=0.87\linewidth]{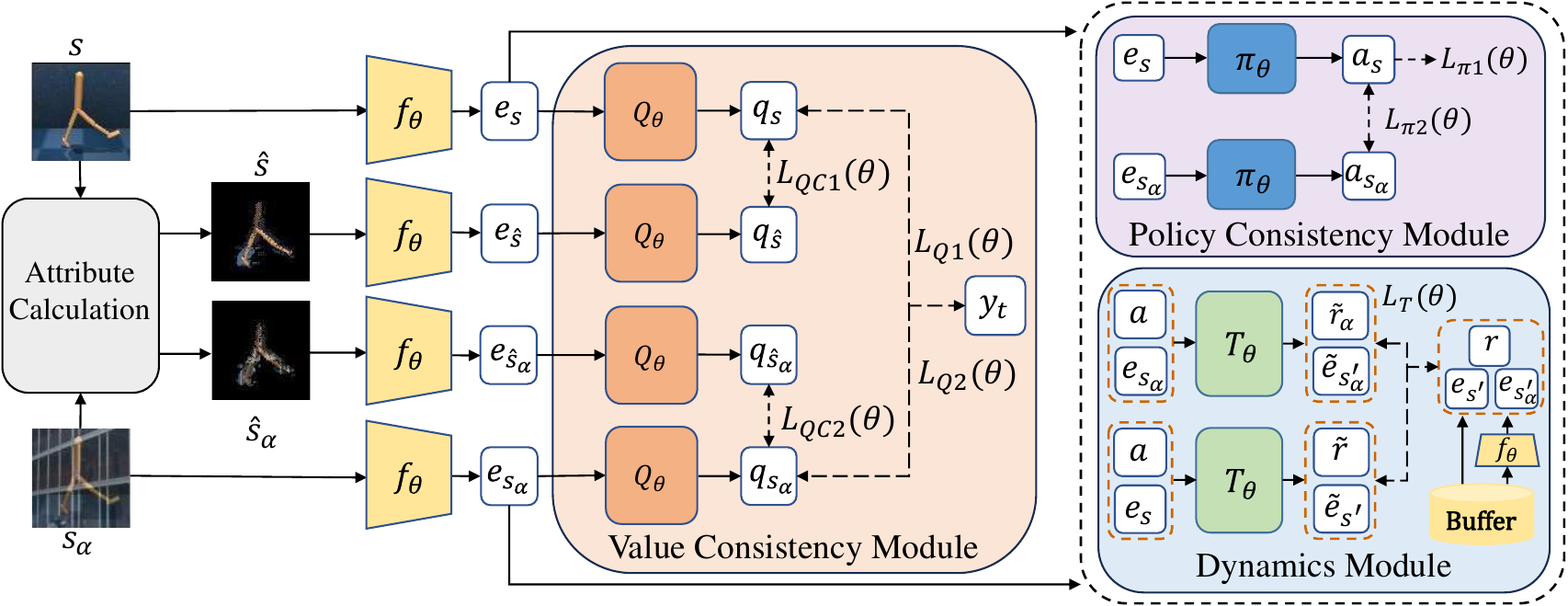}
  \caption{Overview of SCPL. 
  The value consistency module is trained using the original and augmented observations \(s\) and \(s_{\alpha}\), along with their saliency attribute maps \(\hat{s}\) and \(\hat{s}_{\alpha}\). 
  The dynamics module aids the encoder \(f_{\theta}\) in acquiring task-relevant representations, while the policy consistency module introduces a constraint to maintain consistency in action distributions.} 
  \label{fig_all_arch}
  \vspace{-.2cm}
\end{figure*}

\vspace{-.2cm}
\subsection{Representation learning in visual RL} 
Numerous methods \cite{iso-dream, ted} improve generalization by learning task-relevant representations through auxiliary tasks. 
Some approaches \cite{does?} improve representation effectiveness by employing observation reconstruction as auxiliary tasks. 
DBC \cite{dbc} minimizes the bisimulation metric in latent spaces to learn invariant representations without task-irrelevant information. 
PAD \cite{pad} utilizes an inverse dynamic model to predict actions based on current and next states.
Dr.G \cite{drg} trains the encoder and world model using contrastive learning and introduces an inverse dynamics model to capture temporal structure.
These methods learn task-relevant information through the guidance of rewards and dynamic consistency.
However, they struggle to extract task-relevant information for perturbed observations due to their exclusive training on original data and uncritical task-relevant attention.
SCPL achieves task-relevant representations for both original and perturbed observations by training a dynamics module using both original and augmented data while focusing on task-relevant regions.

\vspace{-.2cm}
\subsection{Policy learning for RL} 
Some studies explore the decoupling of value functions and policy networks to learn effective policies or obtain invariant representations \cite{yarats2021improving, andrychowicz2020matters}.
PPG \cite{ppg} mitigates the interference between policy and value optimization by distilling the value function while constraining the policy.
IDAAC \cite{idaac} models both the policy and value function while introducing an auxiliary loss to obtain representations that remain invariant to task-irrelevant properties.
DCPG \cite{dcpg} implicitly penalizes value estimates by optimizing the value network less frequently, using more training data than the policy network.
However, prior studies that focus on learning invariant representations often overlook the consistency of policies across both original and perturbed observations.
In contrast, SCPL learns task-relevant representations while maintaining a consistent superior policy for perturbed observations, similar to that for original observations.
To the best of our knowledge, we are the first to highlight the importance of policy consistency between original and perturbed observations for generalization ability.


\vspace{-.2cm}
\section{PROBLEM FORMULATION} \label{sec3}
Visual reinforcement learning (RL) is considered a partially observable Markov decision process (POMDP) because only partial states are observed from images. A POMDP is defined as a tuple $M=\langle S, O, A, P, r,\gamma \rangle$, where $S$ is the state space, $O$ is the observation space, $A$ is the action space, $P: S \times A \times S \rightarrow \mathbb{R}$ is the transition probability distribution, $r: S \rightarrow \mathbb{R}$ is the reward function, and $\gamma \in (0, 1)$ is the discount factor.
Let $\pi$ denote a stochastic policy and $\eta(\pi)$ denote its expected cumulative reward: $
\eta(\pi)=\mathbb{E}_{s_0, a_0, \ldots}\left[\sum_{t=0}^{\infty} \gamma^t r\left(s_t\right)\right]$
, where $\pi$ is the policy, $a$ is the action, and $s_{t}$ is the state in the $t$ step. 
The purpose of visual RL is to find a policy $\pi^{*}$ to maximize the expected cumulative reward.
The state-action value function $Q_{\pi}$, the value function $V_{\pi}$, and the advantage function $A_{\pi}$ are defined as:
$Q_\pi\left(s_t, a_t\right)=\mathbb{E}_{s_{t+1}, a_{t+1}, \ldots}\left[\sum_{l=0}^{\infty} \gamma^l r\left(s_{t+l}\right)\right]$,
$V_\pi\left(s_t\right)=\mathbb{E}_{a_t, s_{t+1}, \ldots}\left[\sum_{l=0}^{\infty} \gamma^l r\left(s_{t+l}\right)\right]$,
and $A_\pi(s, a)=Q_\pi(s, a)-V_\pi(s)$.

\section{Methodology} \label{sec4}
We propose a salience-invariant consistent policy learning (SCPL) framework to improve the zero-shot generalization of visual RL.
As shown in Fig.\ref{fig_all_arch}, SCPL mainly consists of three modules: the value consistency module, the policy consistency module, and the dynamics module.
In this paper, $\theta$, $\zeta$, $\phi$, and $\psi$ represent the parameters of the encoder, the value function, the policy network, and the dynamic model, respectively.

\subsection{Value Consistency Module}
To extract task-relevant representations from both original and perturbed observations, it is essential for the encoder and value function to consistently focus on task-relevant regions. We introduce a value consistency module with a novel loss function for the encoder and value function, leveraging augmented data and their saliency maps.
To improve the consistency of attention regions for original and perturbed observations, we update the value function with both original and augmented observations. 
The value loss for the original data $L_{Q1}$ is:
\begin{equation}
    L_{Q1}(\theta,\zeta)=\mathbb{E}_{s,a}[(Q_{\zeta}(f_{\theta}(s),a)-y_{t})^{2}]. 
\label{eq1}
\end{equation}
The value loss for the augmented data $L_{Q2}$ is:
\begin{equation}
    L_{Q2}(\theta,\zeta)=\mathbb{E}_{s_{\alpha},a}[(Q_{\zeta}(f_{\theta}(s_{\alpha}),a)-y_{t})^{2}],
\label{eq2}
\end{equation}
where $y_{t}$ represents the target of Q-value. 
These losses encourage the value function to estimate the same values for both original and augmented observations, thereby promoting consistency in the attention regions.
However, maintaining consistency in value estimation alone may be insufficient to ensure that agents focus on task-relevant regions amidst increasing perturbations. 
Therefore, additional guidance is necessary to help agents remain attentive to task-relevant regions in the observations.
\begin{figure}
  \includegraphics[width=.9\linewidth]{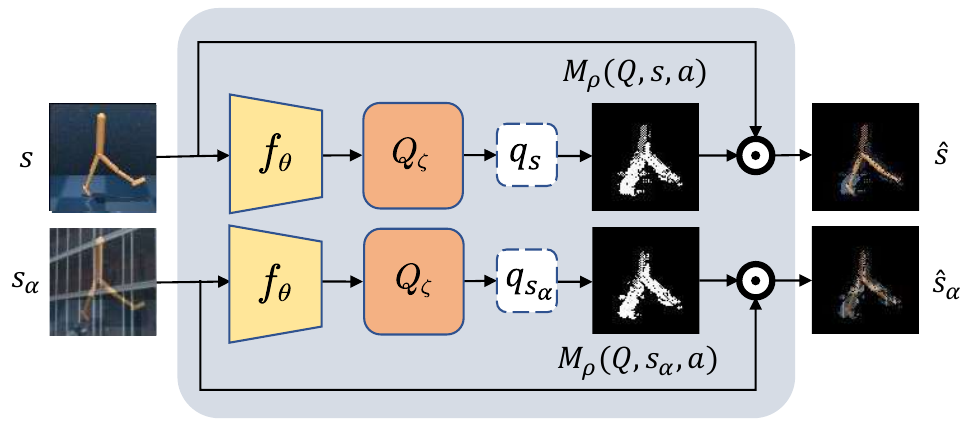}
  \centering
  \caption{The generation of saliency attribute masked maps.} 
  \label{fig_gradient_map}
  \vspace{-.3cm}
\end{figure}

SCPL utilizes saliency attribute masked maps to guide the encoder and the value function to focus on task-relevant regions for observations.
As shown in  Fig.\ref{fig_gradient_map}, we generate the saliency attribute masked maps ($\hat{s}$ and $\hat{s}_{\alpha}$) for original and augmented observations (${s}$ and ${s}_{\alpha}$) using the vanilla gradient method \cite{vanilla}.
We use guided backpropagation to compute the gradient map $M(Q,s,a)$ of the Q-network, represented as $M(Q,s,a)=\partial Q(s,a) / \partial s$. 
Let $M_{\rho}(Q,s,a)$ be the binarized $\rho$-quantile saliency attribute map.
Specifically, if the gradient pixel $M(Q,s,a)_{j}$ belongs to the top $1-\rho$ quantile of gradient values, then $M_{\rho}(Q,s,a)_{j}$ will be set to 1, otherwise 0. 
The $\rho$-quantile saliency attribute map $\hat s$ and $\hat s_{\alpha}$ represents the attention of the value function towards input observations, where white areas indicate regions of high attention and less attended regions are masked out.
Then, the saliency attribute masked maps are generated by multiplying the observations with their saliency attribute maps to show the attended pixels. 
$\bigodot$ denotes the Hadamard product.

To guide agents to focus on task-relevant pixels, we introduce a saliency consistency term between original observations and their respective saliency attribute maps.
The saliency consistency loss for the original data is:
\begin{equation}
     L_{QC1}(\theta,\zeta)=\mathbb{E}_{s,\hat s,a}[(Q_{\zeta}(f_{\theta}(\hat s),a)-Q_{\zeta}(f_{\theta}(s),a))^{2}].
\label{eq3}
\end{equation}
To ensure that agents focus on task-relevant regions in both original and perturbed observations, we extend saliency guidance to augmented data while updating the value function with this augmented data.
The saliency consistency loss for the augmented data is:
\begin{equation}
     L_{QC2}(\theta,\zeta)=\mathbb{E}_{s_{\alpha},\hat s_{\alpha},a}[(Q_{\zeta}(f_{\theta}(\hat s_{\alpha}),a)-Q_{\zeta}(f_{\theta}(s_{\alpha}),a))^{2}].
\label{eq4}
\end{equation}
With the saliency guidance from Eq.(\ref{eq3}) and Eq.(\ref{eq4}), agents are able to focus on task-relevant regions in both original and perturbed observations.

We combine the value loss with the saliency consistency loss to form the training objective. The value function's objective is:
\begin{equation}
\begin{split}
     L_{Q}(\theta,\zeta)=L_{Q1}(\theta,\zeta) + L_{Q2}(\theta,\zeta)
      + \lambda ( L_{QC1}(\theta,\zeta) + L_{QC2}(\theta,\zeta) ),
\end{split}
\end{equation}
where $\lambda$ is the value consistency coefficient.
$L_{Q1}(\theta,\zeta)$ and $L_{Q2}(\theta,\zeta)$ ensure the encoder and  value function attend to consistent regions in both the training and test environments, while $ L_{QC1}(\theta,\zeta)$ and $L_{QC2}(\theta,\zeta)$ ensure agent focus on task-relevant regions within observations.
This value loss enables the encoder and value function to capture consistent task-relevant pixels from both original and perturbed observations.

\subsection{Dynamics Module}
To enable the encoder to effectively provide task-relevant representations, we develop a dynamic model to ensure representations meet the conditions of rewards and dynamics. 
Specifically, we construct this dynamic model by predicting rewards and next-state representations for both the original and augmented observations.
The loss of the dynamics module for embedding $e$ is:
\begin{equation}
     L_{Te}(\theta,\psi) = \mathbb{E}_{s,a}[ (e^{\prime}-P_{\psi}(f_{\theta}(s),a))^{2} + (r-R_{\psi}(f_{\theta}(s),a))^{2} ],
\end{equation}
where $e$ and $e^{\prime}$ are the latent representations of the current observation and the next observation.
Dynamics module $T$ consists of dynamic head $P$ and reward head $R$.
The training objective of the dynamics module of the embedding for augmented data $e_{\alpha}$, is:
\begin{equation}
\begin{split}
    L_{Te_{\alpha}}(\theta,\psi) = \mathbb{E}_{s_{\alpha},a}[(e_{\alpha}^{\prime}-P_{\psi}(f_{\theta}(s_{\alpha}),a))^2 + 
     (r-R_{\psi}(f_{\theta}(s_{\alpha}),a))^2 ].
\end{split}
\end{equation}
The training objective for the dynamics module is:
\begin{equation}
     L_{T}(\theta,\psi) = L_{Te}(\theta,\psi) + L_{Te_{\alpha}}(\theta,\psi).
\end{equation} 
In SCPL, the value consistency module ensures attention to task-relevant regions, while the dynamics module guides representations that align with  reward and dynamics conditions.
With the combination of the value consistency module and the dynamics module, the encoder can generate task-relevant representations.

\subsection{Policy Consistency Module}
As illustrated in Fig.\ref{fig_first} (middle), previous RL agents frequently exhibit poor policy consistency, reflected in the substantial KL divergence, between training and test environments.
In this section, our theoretical analysis reveals that the policy consistency of agents contributes to enhanced generalization capability. Furthermore, we propose a policy consistency module that improves generalization by enhancing agents' policy consistency across both original and perturbed observations.

\textbf{Relationship between policy consistency and generalization ability.}
We utilize the KL divergence of action distributions between training and test environments to assess the policy consistency of agents. 
Additionally, the divergence in cumulative rewards between these environments reflects the agents' generalization capabilities. 
In the context of visual generalization, the training and test environments are identical except for visual observations. 
Consequently, the agent's policies in both environments can be regarded as two equivalent policies within the training environment.
We prove that there is a positive correlation between the upper bound of the divergence in cumulative rewards of the two policies and their KL divergence.

Initially, we utilize the total variation divergence to measure the distance between two policies' distributions. 
The divergence is defined as : $D_{T V}(p \| q)=\frac{1}{2} \sum_i\left|p_i-q_i\right|$ for probability distributions $p$ and $q$. 
Define $D_{\mathrm{TV}}^{\max }(\pi_{o}, \pi_{p})$ as : 
\begin{equation}
D_{\mathrm{TV}}^{\max }(\pi_{o}, \pi_{p})=\max _s D_{T V}(\pi_{o}(\cdot \mid s) \| \pi_{p}(\cdot \mid s)),
\end{equation}
where $\pi_{o}$ and $\pi_{p}$ are policies in training and test environments, respectively. With this measure in hand, we can now state the following theorem:

\setlength{\parindent}{0pt}\textbf{Theorem 1.} \textit{Let $\alpha=D_{\mathrm{TV}}^{\max }\left(\pi_{o}, \pi_{p}\right)$, the following bound holds:}
\begin{equation}
\eta\left(\pi_{o}\right) - \eta\left(\pi_{p}\right) \leq \frac{2 \epsilon \gamma}{(1-\gamma)^2} \alpha^2,
\end{equation}
where $\eta$ is expected return, $\epsilon=\max _{s, a}\left|A_{\pi}(s, a)\right|$. 
According to \cite{pollard}, the relationship between the total variation divergence and the KL divergence is: $D_{T V}(p \| q)^{2} \leq D_{\mathrm{KL}}(p \| q)$. 
Let $D_{\mathrm{KL}}^{\max }(\pi, \tilde{\pi})=\max _s D_{\mathrm{KL}}(\pi(\cdot \mid s) \| \tilde{\pi}(\cdot \mid s))$.
With Theorem 1, the following bound holds:
\begin{equation}
\eta\left(\pi_{o}\right) - \eta\left(\pi_{p}\right)  \leq C D_{\mathrm{KL}}^{\max }(\pi_{o}, \pi_{p}),
\end{equation}
where $C=\frac{2 \epsilon \gamma}{(1-\gamma)^2}$. 
Hence, a smaller KL divergence of action distributions between training and test environments pushes a tighter upper bound on the disparity of cumulative rewards in these environments.
This implies that policy consistency contributes to the generalization performance of agents.

\begin{algorithm}[t]
\caption{SCPL (changes to SAC in \textcolor{blue}{blue})}
\label{alg:algorithm}
\raggedright  
\textbf{Parameter}: $\mathcal{B}$: replay buffer, \textcolor{blue}{$N_{A}$}: dynamics module update frequency, \textcolor{blue}{$\tau$}: data augmentation function, $\alpha$: learning rate, \textcolor{blue}{$\lambda$}: value consistency coefficient, \textcolor{blue}{$\beta$}: policy consistency coefficient.
\vspace{-\baselineskip} 
\begin{algorithmic}[1] 
\FOR{each iteration} 
\do{\STATE Sample a transition: \\
$a, s^{'} \sim \pi_{\phi}(\cdot|s),P(\cdot|s,a)$ 
\STATE Add transition to replay buffer: \\
$\mathcal{B} \leftarrow \mathcal{B} \cup\left\{\left(s, a, \mathcal{R}(s, a), s^{\prime}\right)\right\}$  
\STATE Sample a batch of transition: \\
$\{s, a, r,s^{'}\} \sim \mathcal{B}$
\STATE Generate augmented data: \\
\textcolor{blue}{$s_{\alpha} \leftarrow \tau(s)$}
\STATE Update value consistency module: \\
$\{\theta,\zeta \}\leftarrow \{\theta,\zeta\} - \alpha\nabla_{\{\theta,\zeta\}}(L_{Q1}(\theta,\zeta) \textcolor{blue}{+ \textit{L}_{Q2}(\theta,\zeta) + \lambda L_{QC1}(\theta,\zeta) + \lambda L_{QC2}(\theta,\zeta)})$ 
\STATE Update dynamics module: \\
\textcolor{blue}{$\{\theta,\psi\} \leftarrow \{\theta,\psi\} - \alpha\nabla_{\{\theta,\psi\}}L_{T}(\theta,\psi)$}}
\STATE Update policy consistency module: \\
$\phi \leftarrow \phi - \alpha\nabla_{\phi}(L_{\pi o}(\phi) \textcolor{blue}{+ \beta L_{\pi c}(\phi)})$
\ENDFOR
\end{algorithmic}
\end{algorithm}

\textbf{Policy consistency module.}
To enhance the generalization of visual RL algorithms, agents should produce consistent policies for both original and perturbed observations.
Therefore, we design a policy consistency loss with the KL divergence of action distributions of both the original and augmented data for the policy network.
The policy loss for the original observation is:
\begin{equation}
     L_{\pi 1}(\phi) = \mathbb{E}_{s,a \sim \pi_{\phi}(\cdot|e_{s})}[\alpha \log \pi_{\phi}(a|e_s) - Q(e_s,a) ],
\end{equation}
where $e_{s}$, $e_{s_{\alpha}}$ are embeddings for observations $s$ and $s_{\alpha}$. 
The policy consistency loss is:
\begin{equation}
     L_{\pi 2}(\phi) = \mathbb{E}_{s, s_{\alpha}}[D_{KL}\left( \pi_{\phi}(\cdot|e_s) || \pi_{\phi}(\cdot|e_{s_{\alpha}}) \right) ].
\label{kl}
\end{equation}
With the policy consistency loss, SCPL improves generalization by encouraging agents to generate consistent policies for original and perturbed observations.
In summary, the loss of the policy consistency module is:
\begin{equation}
     L_{\pi}(\phi) = L_{\pi 1}(\phi) + \beta L_{\pi 2}(\phi),
\end{equation}
where $\beta$ is the policy consistency coefficient. 
By minimizing the total loss, the policy consistency module attains a consistently superior policy in test environments, similar to that in training environments.
Algorithm \ref{alg:algorithm} presents the pseudocode for SCPL. 


\vspace{-.1cm}
\section{Experiments} \label{sec5}
In this section, we conduct experiments to investigate the following questions:
(1) Does SCPL exhibit superior visual generalization capability compared to current state-of-the-art methods?
(2) Can SCPL focus on consistent task-relevant pixels in both original and perturbed observations?
(3) Does SCPL possess consistent representations and policies?
(4) What is the contribution of various modules to generalization performance?
(5) Can SCPL demonstrate advanced generalization in challenging robotic and autonomous driving environments?

\vspace{-.1cm}
\subsection{Experimental Settings}
We evaluate the zero-shot generalization performance of our method in DeepMind Control Suite (DMC) \cite{dmc, overlay}, Robotic Manipulation tasks \cite{robot_env}, and CARLA \cite{carla}. 
All methods are trained in the default environment and evaluated with visual perturbations.
In the DMC experiment, we compare the generalization ability of SCPL with SOTA methods including SAC \cite{sac}, SVEA \cite{svea}, SIM \cite{sim}, TLDA \cite{tlda}, PIE-G \cite{pie-g}, SGQN \cite{sgqn}, CG2A \cite{cg2a}, MaDi \cite{madi}, and CNSN \cite{cnsn}. 

\vspace{-.1cm}
\subsection{Evaluation on the DeepMind Control Suite}

We evaluate the agent's generalization ability on five tasks in DMC-GB \cite{overlay}. 
The agent is trained with default backgrounds and evaluated on test environments: \textit{Color hard}, \textit{Video easy}, and \textit{Video hard}. 

\begin{table*}[t]
\caption{DMC-GB Generalization Performance} 
\setlength{\tabcolsep}{5pt}
\centering
\resizebox{2\columnwidth}{!}{
\begin{tabular}{cccccccccccc}
\toprule
 Setting & Task &  SAC\cite{sac} & SVEA\cite{svea} & SIM\cite{sim} & TLDA\cite{tlda} & PIE-G\cite{pie-g} & SGQN\cite{sgqn} & CG2A\cite{cg2a} & MaDi\cite{madi} & CNSN\cite{cnsn} & \textbf{SCPL} \\
\midrule
\multirow{6}{*}{\thead{Color \\ hard} } &
Walker stand &
$423 \pm 155$ &
$942 \pm 26$ &
$940 \pm 2$ &
$947 \pm 26$ &
$941 \pm 35$ &
$948 \pm 25$  & 
$\textbf{972} \pm \textbf{23}$  & 
$-$   & 
$942 \pm 19$  &
$960 \pm 11$ \\
 &
Walker walk &
$255 \pm 61$ &
$760 \pm 145$ &
$803 \pm 33$ &
$823 \pm 58$ &
$884 \pm 20$ &
$810 \pm 43$ & 
$902 \pm 46$  & 
$-$   & 
$815 \pm 65$  &
$\textbf{939} \pm \textbf{19}$    \\
 &
Cartpole &
$615 \pm 29$ &
$837 \pm 23$ &
$841 \pm 13$ &
$760 \pm 60$ &
$749 \pm 46$ &
$806 \pm 6$ & 
$856 \pm 40$  & 
$-$   & 
$679 \pm 35$  &
$\textbf{857} \pm \textbf{12}$   \\
 &
Ball in cup &
$391 \pm 245$ &
$961 \pm 7$ &
$953 \pm 7$ &
$932 \pm 32$ &
$964 \pm 7$ &
$887 \pm 10$ & 
$\textbf{972} \pm \textbf{10}$  & 
$-$   & 
$894 \pm 78$  &
$966 \pm 9$ \\
 &
Finger spin   &
$373 \pm 70$ &
$\textbf{977} \pm \textbf{5}$  &
$960 \pm 6$   &
$-$  &
$-$  &
$899 \pm 27$  & 
$928 \pm 43$  & 
$-$   & 
$-$   & 
$929 \pm 24$   \\ \cline{2-12}
 &
Average &
$411$ &
$895$ &
$899$ &
$865$ &
$884$ &
$870$ &
$926$   & 
$-$   & 
$833$   & 
$\textbf{930(+1\%)}$ \\
\midrule
\multirow{6}{*}{\thead{Video \\ easy} } &
Walker stand &
$351 \pm 245$ &
$961 \pm 8$ &
$963 \pm 5$ &
$\textbf{973} \pm \textbf{6}$ &
$957 \pm 12$ & 
$955 \pm 9$ & 
$968 \pm 6$  & 
$967 \pm 3$  & 
$967 \pm 6$  & 
$968 \pm 8$  \\
 &
Walker walk &
$228 \pm 65$ &
$819 \pm 71$ &
$861 \pm 33$ &
$873 \pm 34$ &
$870 \pm 22$ & 
$910 \pm 24$ &  
$918 \pm 20$ & 
$895 \pm 24$ & 
$842 \pm 58$  & 
$\textbf{941} \pm \textbf{9}$ \\
 &
Cartpole &
$359 \pm 80$ &
$782 \pm 27$ &
$770 \pm 13$ &
$671 \pm 57$ &
$597 \pm 61$ & 
$761 \pm 28$ & 
$788 \pm 24$ & 
$\textbf{848} \pm \textbf{6}$ & 
$752 \pm 26$ & 
$814 \pm 21$  \\
 &
Ball in cup &
$338 \pm 201$ &
$871 \pm 106$ &
$820 \pm 135$ &
$887 \pm 58$ &
$922 \pm 20$ & 
$950 \pm 24$ & 
$\textbf{963} \pm \textbf{28}$ & 
$807 \pm 144$ & 
$913 \pm 45$ & 
$\textbf{963} \pm \textbf{10}$ \\
 &
Finger spin &
$260 \pm 98$ &
$808 \pm 33$ &
$815 \pm 38$ &
$744 \pm 18$ &
$837 \pm 107$ & 
$956 \pm 26$ & 
$912 \pm 69$ & 
$679 \pm 17$ & 
$-$   & 
$\textbf{963} \pm \textbf{8}$ \\ \cline{2-12}
 &
Average &
$300$ &
$848$ &
$845$ &
$830$ &
$837$ &
$906$ &
$909$ & 
$839$ & 
$869$ & 
$\textbf{930(+2\%)}$  \\
\midrule
\multirow{6}{*}{\thead{Video \\ hard} } &
Walker stand &
$225 \pm 58$ &
$747 \pm 43$ &
$827 \pm 24$ &
$602 \pm 51$ &
$852 \pm 56$ &
$851 \pm 24$ & 
$895 \pm 35$ & 
$920 \pm 14$ & 
$871 \pm 23$ & 
$\textbf{953} \pm \textbf{15}$ \\
 &
Walker walk &
$104 \pm 18$ &
$385 \pm 63$ &
$459 \pm 67$ &
$271 \pm 55$ &
$600 \pm 28$ &
$739 \pm 21$ & 
$687 \pm 18$ & 
$504 \pm 33$ & 
$480 \pm 46$ & 
$\textbf{818} \pm \textbf{32}$ \\
 &
Cartpole &
$174 \pm 24$ &
$401 \pm 38$ &
$367 \pm 47$ &
$286 \pm 47$ &
$401 \pm 21$ &
$544 \pm 43$ & 
$472 \pm 24$ & 
$619 \pm 24$ & 
$417 \pm 31$ & 
$\textbf{675} \pm \textbf{3}$  \\
 &
Ball in cup &
$196 \pm 82$ &
$498 \pm 147$ &
$287 \pm 39$ &
$257 \pm 57$ &
$786 \pm 47$ &
$782 \pm 57$ & 
$806 \pm 44$ & 
$758 \pm 135$ & 
$691 \pm 72$ & 
$\textbf{924} \pm \textbf{7}$ \\
 &
Finger spin &
$26 \pm 21$ &
$307 \pm 24$ &
$362 \pm 9$ &
$241 \pm 29$ &
$762 \pm 59$ &
$822 \pm 24$ & 
$819 \pm 38$ & 
$358 \pm 25$ & 
$-$   & 
$\textbf{897} \pm \textbf{22}$ \\ \cline{2-12}
 &
Average &
$145$ &
$467$ &
$460$ &
$331$ &
$680$ &
$747$ &
$736$ & 
$632$ & 
$615$ & 
$\textbf{853(+14\%)}$ \\
\bottomrule
\end{tabular}}
\label{table_dmc_result}
\end{table*}

\begin{figure*}[h]
\includegraphics[width=\linewidth]{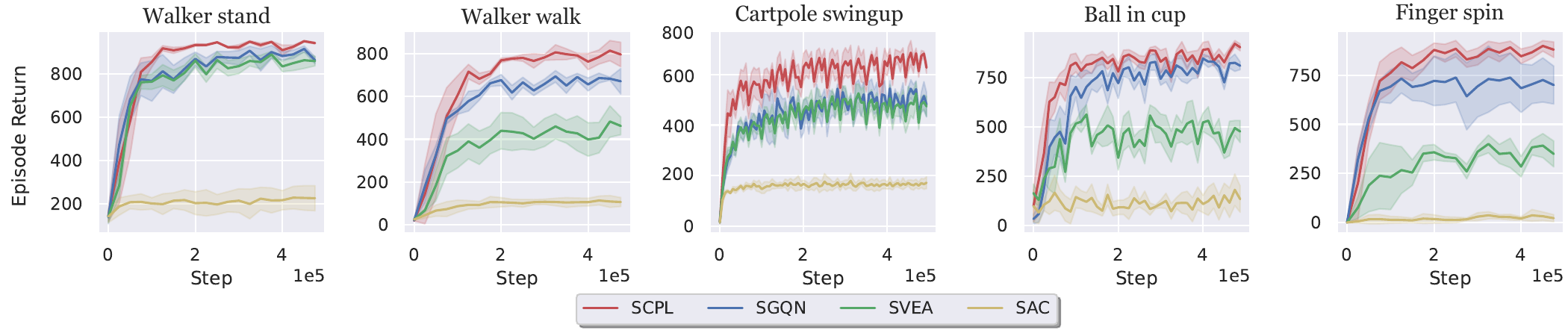}
  \caption{The performance of SAC, SVEA, SGQN, and SCPL in \textit{Video hard} setting. SCPL (red line) shows better generalization.}
  \label{vhard_83}
  \vspace{-.2cm}
\end{figure*}

\textbf{Does SCPL exhibit superior visual generalization
capability?}
We evaluate the visual generalization performance of SCPL across 15 visual perturbed control tasks in the DMC. 
As shown in Table \ref{table_dmc_result}, we report the mean and standard deviation of episode returns over three seeds. 
SCPL agents are trained using two data augmentation techniques from \cite{overlay}: \textit{random convolution} and \textit{random overlay}. 
The SCPL results in Table \ref{table_dmc_result} are based on random convolution for the \textit{Color hard} task, and random overlay for both the \textit{Video easy} and \textit{Video hard} tasks.
Table \ref{table_dmc_result} shows that SCPL outperforms other baselines in 13 out of 15 tasks within unseen test environments. 
Notably, SCPL achieves performance improvements of 12\% in walker stand, 11\% in walker walk, 28\% in cartpole swing-up, 18\% in ball in cup, and 9\% in finger spin tasks in the challenging \textit{video hard} setting.
Overall, SCPL achieves an average performance improvement of 14\% across all tasks in the \textit{video hard} environments. Fig.\ref{vhard_83} presents the test curves for SCPL, SGQN, SVEA, and SAC in these environments, where SCPL demonstrates faster convergence due to its effective task-relevant representations and consistent policies.
The experimental results demonstrate that SCPL exhibits superior visual generalization ability in various perturbed environments.

\begin{figure*}[t]
  \centering
  \includegraphics[width=.85\linewidth]{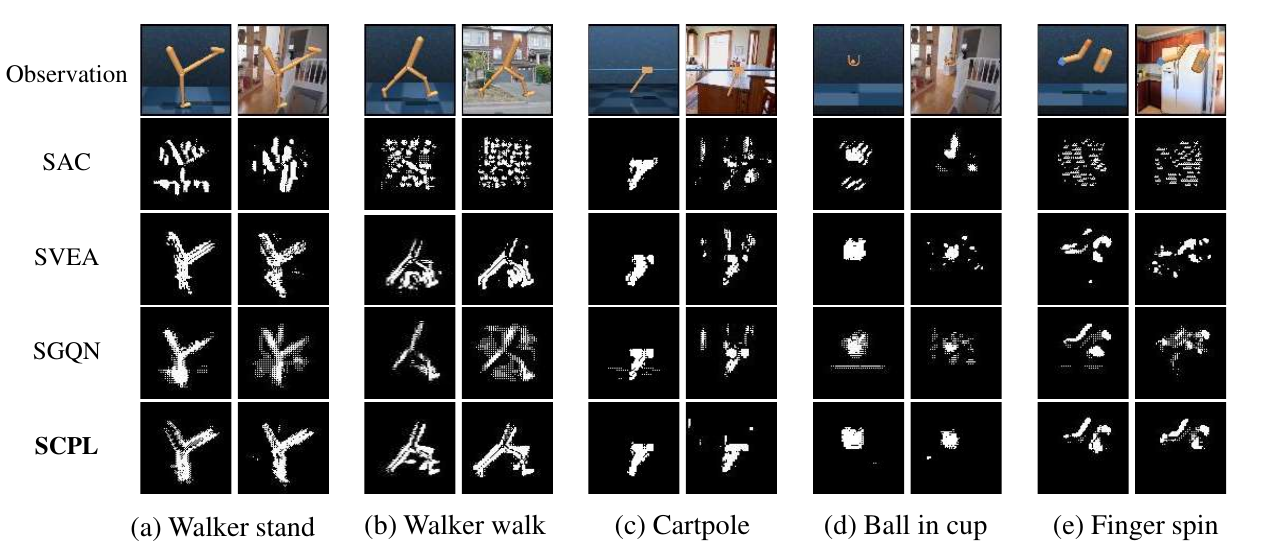}
  \caption{Saliency attribute maps for SAC, SVEA, SGQN, and SCPL in \textit{Training} and \textit{Video hard} setting. In observations of each task, the first column is the original observation, and the second column is the perturbed observation.}
  \label{fig_salience_map2}
  \vspace{-.2cm}
\end{figure*}

\textbf{Can SCPL focus on consistent task-relevant pixels?}
To evaluate the SCPL agent's attention to task-relevant regions, we visualized the saliency maps of agents in both original and perturbed \textit{video hard} observations across five DMC tasks.
Fig.\ref{fig_salience_map2} presents a comparison of the saliency attribute maps for SAC, SVEA, SGQN, and SCPL in the \textit{video hard} setting.
By comparing the saliency maps of the agents in original and perturbed observations, SCPL exhibits similar areas of focus for both types of observations across all tasks, while other baselines typically focus on different regions between the original and perturbed observations.
The comparison indicates that SCPL demonstrates more consistent attention regions.

By comparing the saliency maps of different methods in original and perturbed observations, we find that SCPL consistently focuses on significant task-relevant regions across both types of observations. 
In contrast, other baselines typically capture only approximate task-relevant regions in the original observations and struggle to maintain consistent attention to significant task-relevant areas in the perturbed observations.
Furthermore, Table \ref{analyse} evaluates the accuracy of the agent's attended task-relevant regions in perturbed observations using the following metrics: ACC, measuring overall prediction correctness for pixels; AUC, representing the area under the Receiver Operating Characteristic (ROC) curve; and F1 score, considering both precision and recall to compute a unified score.
These metrics are averaged across five \textit{Video hard} tasks within the DMC tasks.
According to the comparison of saliency maps and statistical results, it's evident that SCPL captures more critical task-relevant regions within various perturbed observations compared to the other methods.
Thus, SCPL can consistently focus on task-relevant pixels in both original and perturbed observations.
\begin{table}[h]
\vspace{-.2cm}
\caption{ Metrics for attention region of RL agents}
\centering
\resizebox{0.6\linewidth}{!}{
\begin{tabular}{lllll}
\toprule
 &SAC  &SVEA  &SGQN  &\textbf{SCPL}  \\
\midrule
 ACC &0.889  &0.926  &0.932 &\textbf{0.942}\\
 AUC &0.811  &0.833  &0.862 &\textbf{0.908} \\
 F1 &0.341   &0.462  &0.463 &\textbf{0.566}  \\
\bottomrule
\end{tabular}
}
\label{analyse}
\vspace{-.2cm}
\end{table}

\textbf{Does SCPL possess consistent representations and policies?}
\begin{figure*}[t]
\centering
  \includegraphics[width=.8\linewidth]{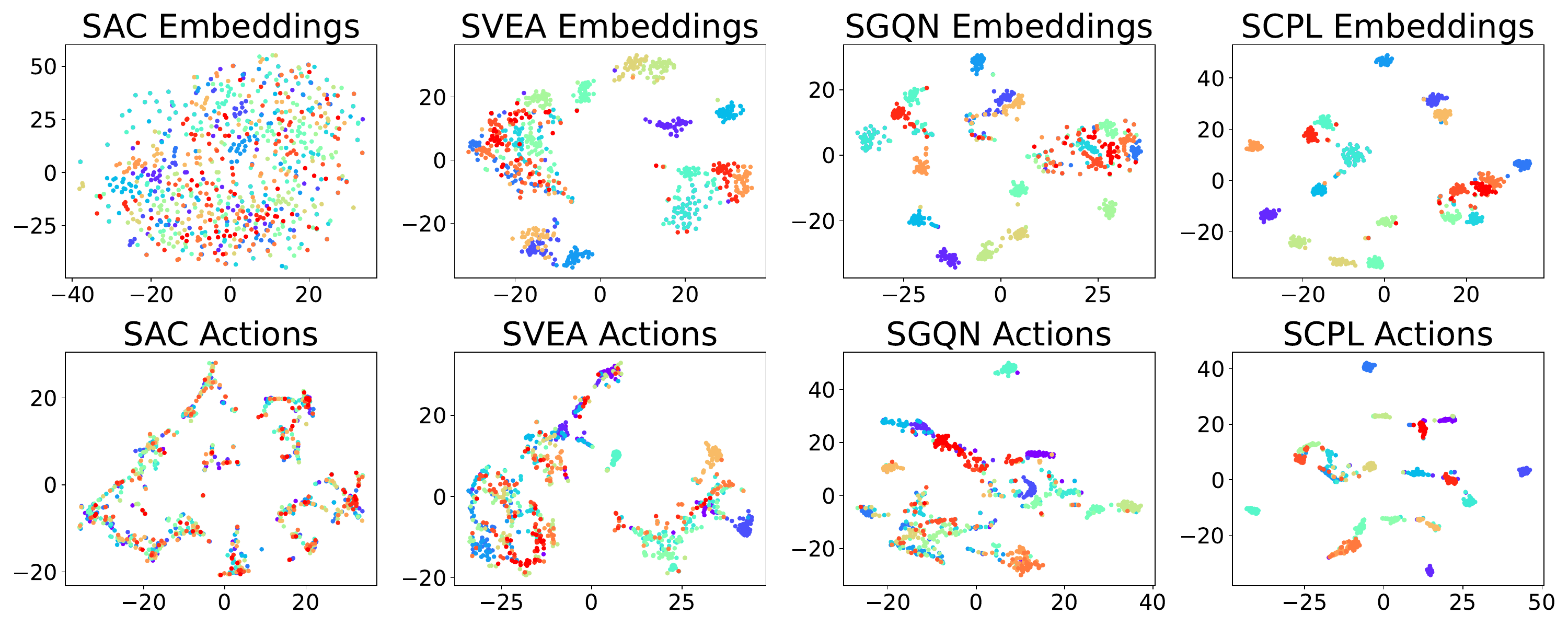}
  \caption{t-SNE maps of embeddings and actions learned with SVEA, SGQN, and SCPL for 20 motion situations, generated by randomly selecting 40 backgrounds from \textit{Video hard}. Different motion situations are represented by different colors, and dots represent representations or actions.}
  \label{fig_tsne}
  \vspace{-.3cm}
\end{figure*}
To further evaluate the task relevance of SCPL's representations and the consistency of its policies, we employed principal component analysis (PCA) to project the representations and actions onto a two-dimensional plane.
We plotted the t-SNE visualization of the agent's representations and actions for 800 observations, composed of 20 motions, each featuring various \textit{video hard} backgrounds. 
Dots of the same color represent representations or actions corresponding to observations with the same motion but different backgrounds.

The first row of Fig. \ref{fig_tsne} displays the t-SNE maps of embeddings learned using SAC, SVEA, SGQN, and SCPL. 
In the t-SNE map for SCPL, dots of the same color cluster closely together, while clusters of different colors are distinctly separated compared to baseline methods. 
This demonstrates that SCPL generates consistent task-relevant representations for perturbed observations, similar to those for original observations.
The second row of Fig. \ref{fig_tsne} presents t-SNE maps of actions in perturbed observations. In the t-SNE maps of baseline methods, actions for observations with different motions cluster together, indicating inconsistency in their policies. In contrast, actions of SCPL for various perturbed observations with the same motion tend to cluster together, while actions for observations with different motions are clearly separated. The t-SNE maps indicate that SCPL is capable of generating consistent task-relevant representations and policies.

\vspace{-.3cm}
\subsection{Ablation Study}

\begin{table*}[t]
\caption{Ablation study of three significant components in SCPL }
\centering
\small
\resizebox{.9\linewidth}{!}{
\begin{tabular}{cccccccc}
\toprule
Benchmark & Environment & SAC  & \thead{\normalsize SAC + \\ \normalsize dynamics module} & \thead{\normalsize SAC + \\ \normalsize value consistency} & \thead{\normalsize SAC + value +\\ \normalsize policy consistency } & \textbf{SCPL} \\

\midrule
\multirow{6}{*}{Video hard } &
Walker stand &
$225 \pm 58$ &
$630 \pm 26$ &
$918 \pm 29$ &
$949 \pm 9$ &
$\textbf{953} \pm \textbf{15}$ \\
 &
Walker walk &
$104 \pm 18$ &
$336 \pm 14$ &
$673 \pm 9$ &
$812 \pm 20$ &
$\textbf{818} \pm \textbf{32}$ \\
 &
Cartpole &
$174 \pm 24$ &
$351 \pm 15$ &
$567 \pm 64$ &
$624 \pm 52$ &
$\textbf{675} \pm \textbf{3}$ \\
 &
Ball in cup &
$196 \pm 82$ &
$405 \pm 29$ &
$805 \pm 67$ &
$909 \pm 12$ &
$\textbf{924} \pm \textbf{7}$  \\
 &
Finger spin &
$26 \pm 21$ &
$292 \pm 6$ &
$703 \pm 15$ &
$802 \pm 6$ &
$\textbf{897} \pm \textbf{22}$ \\  \cline{2-7}
 &
Average &
$145$ &
$403(+\textbf{178}\%)$ &
$733(+\textbf{405}\%)$ &
$819(+\textbf{465}\%)$ &
$\textbf{853}(+\textbf{488}\%)$ \\
\bottomrule
\end{tabular}
}
\vspace{-.1cm}
\label{table4}
\end{table*}

\textbf{What is the contribution of various modules?} 
SCPL leverages the value consistency module, policy consistency module, and dynamics module to enhance generalization. 
To assess the contribution of each component, we evaluate the generalization performance of SAC with the different modules and analyze their respective and combined effects.
The results are demonstrated in Table \ref{table4}.
\textit{SAC + dynamics module} and \textit{SAC + value consistency} refer to the application of the dynamics module and the value consistency module to SAC, respectively.
\textit{SAC + value + policy consistency} represents applying both the value consistency module and the policy consistency module to SAC.
The percentages denote the enhanced performance of modules within SCPL compared to the performance of vanilla SAC.
Specifically, the dynamics module yields improvements of 101\% in \textit{color hard} environments, 166\% in \textit{video easy}, and 178\% in \textit{video hard} environments. The value consistency module achieves impressive gains of 97\% in \textit{color hard}, 191\% in \textit{video easy}, and 405\% in \textit{video hard}. When combining both the value and policy consistency modules, generalization improves further, resulting in performance gains of 112\%, 206\%, and 465\% across the three environments, respectively. In SCPL, the integration of these modules significantly boosts performance across all modes. The ablation results demonstrate that each component plays a crucial role in enhancing SCPL's visual generalization.

\subsection{Generalization in robotic and autonomous driving environments?}
\textbf{Evaluation on Vision-based Robotic Manipulation.}
To further evaluate the generalization ability of the proposed SCPL, we consider three robot manipulation tasks based on third-person visual input introduced in \cite{robot_env}: \textit{Reach}, \textit{Push}, and \textit{Peg in Box}. All agents are trained using the default settings and evaluated in two modes. The easy mode substitutes the default environment with five different background colors and desktop textures, while the hard mode further replaces the desktop textures with complex images. 
We compare SCPL with baseline algorithms SAC, SVEA, and SGQN. The results, presented in Table \ref{table_robot}, demonstrate that SCPL outperforms the best prior methods in terms of generalization, achieving an average improvement of +7\% on the training set, +52\% on the easy set, and +39\% on the hard set. 
The experimental results indicate that SCPL outperforms previous methods in robotic environments.

\begin{table}[H]
\vspace{-.2cm}
\caption{Performance comparison on Robotic Manipulation }
\centering
\resizebox{\linewidth}{!}{
\begin{tabular}{cccccc}
\toprule
 Setting & Task & SAC & SVEA & SGQN & \textbf{SCPL(ours)} \\
\midrule
\multirow{4}{*}{Train} &
Reach &
$1.5   \pm 6.7 $ &
$33.6  \pm 0.6 $ &
$33.6  \pm 0.7 $ &
$\textbf{33.8} \pm \textbf{0.3}$  \\
&
Push &
$-25.3   \pm 13.4  $ &
$10.8    \pm 7.0  $ &
$18.8    \pm 6.4 $ &
$\textbf{19.2} \pm \textbf{6.1} $   \\
&
Peg &
$-12.6  \pm 13.2  $  &
$152.6  \pm 21.1  $  &
$179.8  \pm 23.1  $  &
$\textbf{194.6}   \pm \textbf{12.1}$ \\ \cline{2-6}
 &
Average &
$-12.1$ &
$65.7$ &
$77.4$ &
$\textbf{82.6(+7\%)}$  \\

\midrule
\multirow{4}{*}{ Test easy} &
Reach &
$-22.7   \pm 6.4 $ &
$32.2   \pm 1.0  $ &
$28.2   \pm 5.9  $ &
$\textbf{33.3 } \pm \textbf{0.3}$   \\
&
Push &
$-23.6    \pm 11.2  $ &
$2.9      \pm 10.4   $ &
$-12.6    \pm 12.6   $ &
$\textbf{6.4}    \pm \textbf{6.5}   $   \\
&
Peg &
$-33.6     \pm 20.2  $  &
$110.4     \pm 44.3   $  &
$94.6      \pm 12.0   $  &
$\textbf{181.0} \pm \textbf{14.0}$  \\ \cline{2-6}
&
Average &
$-26.6$ &
$48.5$ &
$36.7$ &
$\textbf{73.6(+52\%)}$  \\

\midrule
\multirow{4}{*}{Test hard} &
Reach &
$-19.9  \pm 4.8 $ &
$27.8  \pm 1.2 $ &
$18.9  \pm 4.6 $ &
$\textbf{31.9 } \pm \textbf{2.0}$   \\
&
Push &
$-24.2   \pm 11.0  $ &
$\textbf{-1.7}   \pm \textbf{13.5}  $ & 
$-17.7   \pm 11.3  $ &
$-3.1  \pm 5.1 $   \\
&
Peg &
$-25.1   \pm 5.2  $  &
$114.0   \pm 43.6  $  &
$124.8   \pm 28.8  $  &
$\textbf{166.4} \pm \textbf{15.0}$ \\ \cline{2-6}
 &
Average &
$-23.1$ &
$46.7$ &
$42.0$ &
$\textbf{65.1(+39\%)}$  \\

\bottomrule
\end{tabular}
}
\label{table_robot}
\vspace{-.2cm}
\end{table}

\textbf{Evaluation on CARLA autonomous driving environments.}
CARLA \cite{carla} is a widely used simulator for autonomous driving. In our generalization experiments \cite{secant}, the agents aim to navigate along the road in the \textit{Highway Town04} map, striving to travel as far as possible without colliding within 1000 time steps. The agent is trained under clear noon weather conditions and evaluated in five different weather scenarios, which include varying lighting conditions, realistic rain, and slippery surfaces. We adapted the reward function to align with the settings used in prior work \cite{dbc}. 

In Table \ref{table_carla}, we present the average driven distance without collisions for vehicles across different weather conditions. Averaged over 10 episodes per weather condition and three training runs, SCPL is able to drive, on average, 69\% farther than previous baselines during tests. Notably, in the \textit{sunset} weather scenario, where all other methods struggle, SCPL demonstrates exceptional generalization capabilities. These experimental results indicate that SCPL achieves superior visual generalization performance in CARLA's autonomous driving environments.

\begin{table}[H]
\vspace{-.2cm}
\caption{Performance comparison on CARLA}
\setlength{\tabcolsep}{2pt} 
\centering
\resizebox{.9\linewidth}{!}{
\begin{tabular}{ccccc}
\toprule
 Setting & SAC & SVEA & SGQN & \textbf{SCPL(ours)} \\
\midrule
Train &
$472  \pm 110  $ &
$297   \pm 14  $ &
$614  \pm 41 $ &
$\textbf{643}   \pm \textbf{87}  $  \\
Wet noon &
$468  \pm 68 $ &
$353  \pm  112 $ &
$473  \pm 187 $ &
$\textbf{564}   \pm \textbf{123}  $  \\
Hard rain noon &
$306  \pm 114 $ &
$268 \pm  89 $ &
$406  \pm 63 $ &
$\textbf{442}   \pm \textbf{199}  $  \\
Wet sunset &
$23  \pm 16 $ &
$125    \pm  36   $ &
$39  \pm 17 $ &
$\textbf{271}  \pm  \textbf{28} $  \\
Soft rain sunset &
$45 \pm 25$ &
$22  \pm   5$  &
$59 \pm 44$ &
$\textbf{243}  \pm \textbf{29} $ \\
Mid rain sunset &
$44 \pm 24  $ &
$42   \pm  31 $  &
$63 \pm 46 $ &
$\textbf{242}  \pm \textbf{11} $ \\
\midrule
Test Average &
$177 $ &
$162 $  &
$208 $ &
$\textbf{352(+69\%)} $ \\
\bottomrule
\end{tabular}
}
\label{table_carla}
\vspace{-.2cm}
\end{table}

\section{Conclusion} \label{sec6}
This paper proposes a Salience-Invariant Consistent Policy Learning (SCPL) algorithm for generalization in visual RL.
SCPL improves visual generalization by promoting task-relevant representations through its value consistency module, which ensures consistent focus on critical regions in both original and perturbed observations, and its dynamics module, which learns dynamics-relevant features.
Additionally, our theoretical analysis reveals that maintaining policy consistency between original and perturbed observations is crucial for visual generalization. 
Therefore, we propose a policy consistency module to enhance generalization performance by reinforcing policy consistency.
Through the extensive experiment results, SCPL demonstrates superior zero-shot generalization performance compared to prior SOTA methods.
In this study, SCPL employs fixed saliency quantiles during training. Exploring adaptive quantiles for saliency maps to enhance task-relevant attention regions presents a promising direction for future research.



\begin{acks}
This work is supported by the National Key Research and Development Program of China under Grants 2022YFA1004000, the Beijing Natural Science Foundation under No. 4242052, the National Natural Science Foundation of China under Grants 62173325, and the CAS for Grand Challenges under Grants 104GJHZ2022013GC.
\end{acks}



\bibliographystyle{ACM-Reference-Format} 
\bibliography{sample}


\clearpage
\onecolumn
\appendix
\section*{Appendix}

\section{Architecture Overview}
\begin{figure*}[!hb]
  \centering
  \includegraphics[width=.9\linewidth]{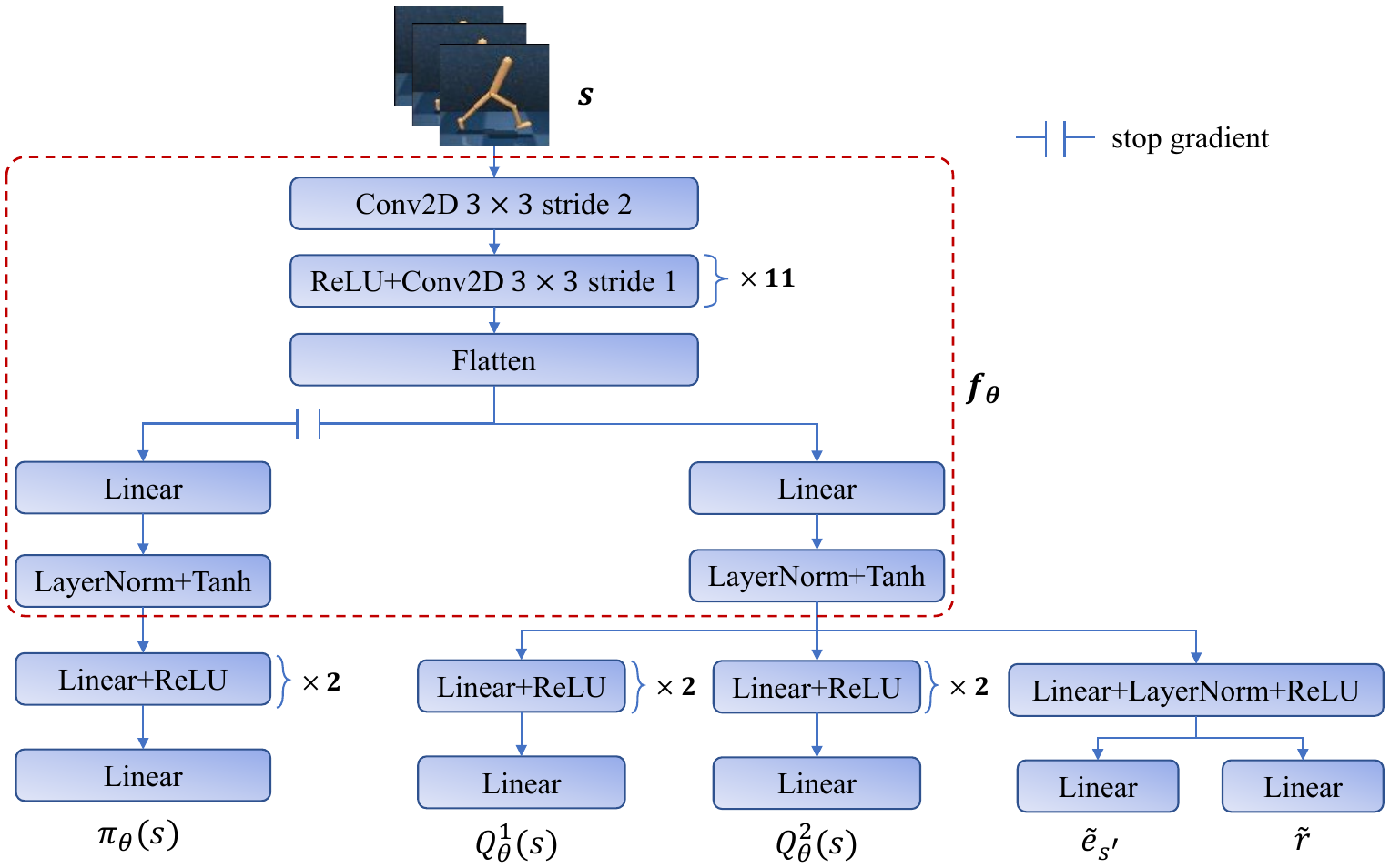}
  \caption{The network architecture of SCPL agent.}
  \label{fig_architecture}
\end{figure*}

We implement the SCPL within the SAC framework. The network architecture of SCPL is depicted in Fig.\ref{fig_architecture}. SCPL comprises three main components: the value module, the policy module, and the dynamics module. For simplicity, we use $\theta$ to represent the parameters of all the networks.

The value module can be divided into two primary components: an encoder $f_{\theta}$ and a value function $Q_{\theta}$. 
The encoder $f_{\theta}$ is composed of a convolutional neural network with 11 convolutional layers and 32 convolutional kernels.
On the other hand, the value function $Q_{\theta}(s)$ is constructed with three fully connected layers, each comprising 1024 neurons. 
For a given input observation $s$, the state-action value can be decomposed as $Q = Q_{\theta}(f_{\theta}(s), a)$.
During the training of the value network, both the encoder $f_{\theta}$ and the value function $Q_{\theta}$ are updated.
The policy network also consists of an encoder $f_{\theta}$ and a policy function $\pi_{\theta}$. 
The encoders in the value module, policy module, and dynamics module share identical parameters. 
The action performed by the policy network for a given input observation $s$ can be expressed as $a = \pi_{\theta}(f_{\theta}(s))$. 
Due to the mutual influence during the training of the policy and value networks \cite{ppg}, \textbf{when updating the policy network, only the policy function $\pi_{\theta}$ is modified, while the encoder $f_{\theta}$ remains unchanged. }
The policy function $\pi_{\theta}$ is also composed of a 3-layer fully connected network. 
The dynamics module consists of an encoder $f_{\theta}$ and a dynamics network $T_{\theta}$, with the encoder sharing parameters with the policy and value networks.
The dynamics module generates the estimated representation $\tilde{e}_{s^{\prime}} $ and estimates reward $\tilde{r}$ based on the current state $e_{s}$ and the action $a$. 
It can be represented as $\tilde{e}_{s^{\prime}}, \tilde{r} = T_{\theta}(f_{\theta}(s), a)$. 
The dynamic model is employed as an auxiliary task to encourage encoders to provide robust latent representations for the value function and the policy network. Thus, \textbf{both the dynamics network and the encoder are updated during training}. 
The dynamics network comprises three fully connected layers.

In order to capture temporal information, we employ a sequence of three consecutive frames as input. 
This stacking of images is a common approach in reinforcement learning. 
Lastly, the architecture of the agent in the SCPL framework is visualized in Fig.\ref{fig_architecture}.

\clearpage
\section{Proof of the Relationship Between Cumulative Rewards Difference and Policy Distance \label{proof}}
Theorem 1 demonstrates a positive relationship between the upper bound of the difference in cumulative rewards for two policies and the distance between their action distributions. The following is the proof for Theorem 1.
Our proof relies on the concept of coupling, where we define two policies, $\pi_{o}$ and $\pi_{p}$, which take the same actions with a high probability of $1-\alpha$.
As a result, the disparity in expected cumulative rewards between the two policies depends on the varying choices made by these policies.

To prove Theorem 1, we begin by proving Lemma 1, which demonstrates that the difference of two policies' expected discounted reward $\eta(\pi_{o})- \eta(\pi_{p})$ can be decomposed as the sum of the per-timestep advantage over the episode.

\noindent  \textbf{Lemma 1.} \cite{aoarl} \textit{Given two policies $\pi_{o}$ and $\pi_{p}$, the difference of two policies' performance is:}
\begin{equation}
\eta(\pi_{o}) - \eta(\pi_{p}) = \mathbb{E}_{\tau \sim \pi_o}\left[\sum_{t=0}^{\infty } \gamma^t A_{\pi_p}(s_t,a_t)\right].
\end{equation}
This expectation is taken over trajectories $\tau := (s_0,a_0,s_1,a_1,\cdots)$, and $\tau \sim \pi$ indicates that actions are taken from $\pi$ to generate $\tau$.

\noindent  \textbf{\textit{Proof.}}
Note that the definition of the advantage function $A_{\pi_p}(s_t,a_t) = V_{\pi_p}(s_{t+1})- V_{\pi_p}(s_{t})$, hence the following equation holds:
\begin{equation}
    \begin{aligned}
\eta(\pi_{o}) - \eta(\pi_{p}) &= 
\mathbb{E}_{\tau \sim \pi_{o}}\left[\sum_{t=0}^{\infty } \gamma^tr(s_t)\right] - \mathbb{E}_{s_{0}\sim \pi_{p}}\left[V_{\pi_p}(s_0)\right] \\
& = \mathbb{E}_{\tau \sim \pi_{o}}\left[-V_{\pi_p}(s_0)+\sum_{t=0}^{\infty } \gamma^tr(s_t)\right] \\
& = \mathbb{E}_{\tau \sim \pi_{o}}\left[\sum_{t=0}^{\infty }\gamma^t (r(s_t)+\gamma V_{\pi_p}(s_{t+1})- V_{\pi_p}(s_{t}) ) \right] \\
& = \mathbb{E}_{\tau \sim \pi_{o}}\left[\sum_{t=0}^{\infty } \gamma^t A_{\pi_p}(s_t,a_t)\right]. \\
    \end{aligned}
\end{equation}

To prove the relationship between the disparity in cumulative rewards and the distance between two policies, it is essential to select a suitable measurement approach.
We measure the distance between two policies by the probability that the two policies take identical actions at each time step.
The polices $\pi_{o}$ and $\pi_{p}$ denote the marginal distributions of $a_o$ and $a_p$, respectively.

\noindent  \textbf{Definition 1.} \textit{A policy pair $(\pi_{o}, \pi_{p})$ is $\alpha-$coupled if for each action pair $(a_{o}, a_{p})$ sampled from the policy pair $(\pi_{o}, \pi_{p})$, it holds that $P(a_{o} \neq a_{p}) \leq \alpha$ for all states $s$.}

The concept of $\alpha-$coupling implies that, for the same seed and input, the sampled actions of policies $\pi_{o}$ and $\pi_{p}$ agree with the probability of at least $1-\alpha$.

\noindent  \textbf{Lemma 2.}\cite{trpo} \textit{If policies $\pi_{o}$ and $\pi_{p}$ are $\alpha-$coupled, then for all states $s$:}
\begin{equation}
\left|\mathbb{E}_{a \sim \pi_{o}(\cdot|s)}\left[A_{\pi_p}(s,a)\right]\right| \le 2 \alpha \max_{s,a} |A_{\pi_p}(s,a)|.
\label{eqt_lemma2}
\end{equation}
\noindent  \textbf{\textit{Proof.} }
Since $\mathbb{E}_{a \sim \pi}\left[A_{\pi}(s,a)\right] = \mathbb{E}_{a \sim \pi}\left[Q_{\pi}(s,a) - V_{\pi}(s)\right] = 0$, then 
\begin{equation}
    \begin{aligned}
\mathbb{E}_{a \sim \pi_{o}(\cdot|s)}\left[A_{\pi_p}(s,a)\right]
&= \mathbb{E}_{(a_p,a_o) \sim (\pi_{p},\pi_{o})}\left[A_{\pi_p}(s,a_o) - A_{\pi_p}(s,a_p)\right] \\
&=P(a_p \ne a_o | s) \mathbb{E}_{(a_p,a_o) \sim (\pi_{p},\pi_{o})}\left[A_{\pi_p}(s,a_o) - A_{\pi_p}(s,a_p)\right]  \\
&\le 2 \alpha \mathbb{E}_{(a_p,a_o) \sim (\pi_{p},\pi_{o})}\left[A_{\pi_p}(s,a_o) - A_{\pi_p}(s,a_p)\right].
    \end{aligned}
\end{equation}
Taking absolute values on both sides, (\ref{eqt_lemma2}) holds.

\noindent  \textbf{Theorem 1.} \textit{Let $\alpha=D_{\mathrm{TV}}^{\max }\left(\pi_{o}, \pi_{p}\right)$, the following bound holds:}
\begin{equation}
\eta\left(\pi_{o}\right) - \eta\left(\pi_{p}\right) \leq \frac{2 \epsilon \gamma}{(1-\gamma)^2} \alpha^2,
\label{theorem1_1}
\end{equation}
\textit{where $\epsilon=\max _{s, a}\left|A_{\pi}(s, a)\right|$. }

\noindent  \textbf{\textit{Proof.} }
Let $n_t$ denote the number of times that $a_i \neq a_o$ for $i<t$, then the expected discount cumulative rewards can be divided into two parts with $P(n_t=0)$ and $P(n_t>0)$. 

\begin{equation}
\eta(\pi_o) =P(n_t=0) \mathbb{E}_{s_t \sim \pi_o | n_t=0}\left[\sum_{t=0}^{\infty} \gamma^t r\left(s_t\right)\right] + P(n_t>0) \mathbb{E}_{s_t \sim \pi_o | n_t>0}\left[\sum_{t=0}^{\infty} \gamma^t r\left(s_t\right)\right].
\end{equation}
A similar decomposition is applied to the accumulated rewards for actions sampled using policy $\pi_p$:
\begin{equation}
\eta(\pi_p) =P(n_t=0) \mathbb{E}_{s_t \sim \pi_p | n_t=0}\left[\sum_{t=0}^{\infty} \gamma^t r\left(s_t\right)\right] + P(n_t>0) \mathbb{E}_{s_0 \sim \pi_o | n_t>0}\left[V_{\pi_p}(s_{0})\right].
\label{19}
\end{equation}
Note that the $n_t = 0$ terms are equal:
\begin{equation}
\mathbb{E}_{s_t \sim \pi_o | n_t=0}\left[\sum_{t=0}^{\infty} \gamma^t r\left(s_t\right)\right] = \mathbb{E}_{s_t \sim \pi_p | n_t=0}\left[\sum_{t=0}^{\infty} \gamma^t r\left(s_t\right)\right],
\label{20}
\end{equation}
because $n_t < 0$ implies that policies $\pi_o$ and $\pi_p$ sampled the same actions for all time-steps less than t.

By the definition of $\alpha$, we get $P(n_t=0) \ge (1-\alpha)^t $ and  $P(n_t > 0) \le 1 - (1-\alpha)^t $.
According to Lemma 2, defining $\epsilon = \max_{s,a} |A_{\pi_p}(s,a)|$,the left of (\ref{theorem1_1}) can be further simplified as:
\begin{equation}
\begin{aligned}
    \eta(\pi_{o}) - \eta(\pi_{p}) & = P(n_t>0) \left[\mathbb{E}_{\tau \sim \pi_{o}}\left[\sum_{t=0}^{\infty } \gamma^tr(s_t)\right] - \mathbb{E}_{s_{0}\sim \pi_{p}}\left[V_{\pi_p}(s_0)\right]\right] \\
    & = P(n_t>0)\mathbb{E}_{\tau \sim \pi_{o} | n_t>0}\left[\sum_{t=0}^{\infty } \gamma^t A_{\pi_p}(s_t,a_t)\right] \\
    & \le \sum_{t=0}^{\infty } 2 \alpha \epsilon \gamma^t (1-(1-\alpha)^t) \\ 
    & = 2 \alpha\epsilon  \left(\frac{1}{1-\gamma} - \frac{1}{1-\gamma(1-\alpha)}\right)\\
    & = \frac{2\alpha^2\gamma\epsilon}{(1-\gamma)(1-\gamma(1-\alpha))}\\
    & \le \frac{2\alpha^2\gamma\epsilon}{(1-\gamma)^2}.
\end{aligned}
\end{equation}

\clearpage
\section{Baseline Methods \label{baseline}}
In the visual generalization task, we compared SCPL with (i) a pure reinforcement learning algorithm, SAC \cite{sac}, and (ii) reinforcement learning algorithms designed for generalization with data augmentation, including SVEA \cite{svea}, SIM \cite{sim}, TLDA \cite{tlda}, SGQN \cite{sgqn}, CG2A \cite{cg2a}, CNSN\cite{cnsn}, and MaDi \cite{madi}.
The detailed baseline methods are presented below:
\begin{itemize} 
\item SAC \cite{sac}: SAC presents a soft actor-critic algorithm without augmentation.
\item SVEA \cite{svea}: SVEA improves generalization by updating the value function with both original and augmented data. It advocates updating the value function of the augmented data with the target value of the original data to improve the stability of the algorithm.
\item SIM \cite{sim}: SIM designs a cross-correlation matrix for self-supervised representation learning and utilizes the redundancy reduction-based self-supervised loss as an intrinsic reward.
\item TLDA \cite{tlda}: TLDA identifies task-correlated pixels with large Lipschitz constants and selectively augments task-irrelevant pixels to guide agents in learning relevant information. 
\item PIE-G \cite{pie-g}: PIE-G learns the representations with a pre-trained image encoder to improve visual generalization; 
\item SGQN \cite{sgqn}: SGQN employs saliency guidance to direct agents' attention to task-relevant areas in original observations. 
It further aligns agents’ attention across original and augmented data with a trainable network.
However, as depicted in Fig.\ref{fig_first}, the network results in ambiguous attention regions for augmented data, containing more task-irrelevant information compared to the original data.
SGQN captures task-relevant regions in original observations, but the alignment network limits its attention to perturbed observations.
\item CG2A \cite{cg2a}: CG2A proposes a general policy gradient optimization framework to adaptively balance the varying gradient magnitudes and alleviate the gradient conflict bias caused by data augmentation.
\item CNSN \cite{cnsn}: CNSN employs normalization techniques to improve visual generalization in reinforcement learning.
\item MaDi\cite{madi}:MaDi proposes that identifying task-relevant areas helps improve the generalization ability of agents. 
It incorporates a mask network before the value function to mask out task-irrelevant regions in the input images. 
The mask network is updated using value loss, which enhances the agent's focus on task-relevant areas.
\end{itemize} 
The aforementioned data augmentation methods focus on task-relevant regions in original observations but struggle to capture task-relevant pixels for perturbed observations with a lack of guidance.
In our work, agents focus on task-relevant regions in both the original and perturbed data by updating the value function with augmented data and their corresponding saliency maps.

\section{Data Augmentation \label{data_aug}}

\begin{figure*}[!hbt]
\centering
  \includegraphics[width=.65\linewidth]{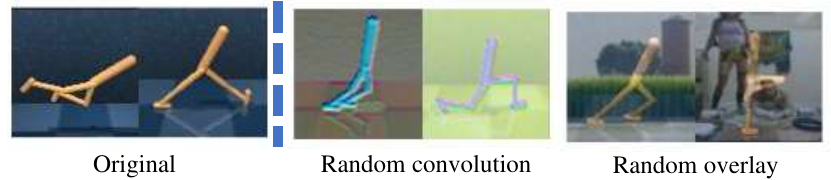}
  \caption{Samples from two data augmentations.}
  \label{fig_da}
\end{figure*}

We consider two data augmentation methods in this paper, viz. random convolution and random overlay, abbreviated as (\textit{conv} and \textit{overlay}), which were shown to be valid in \cite{tlda} and \cite{sim}. 
Random convolution \cite{random_conv}: Change the color of an image by passing it through a random convolution layer.
Random overlay \cite{overlay}: Combine the original observation $s$ with another image $\hat{s}$ using linear interpolation. Specifically, the augmented observation $s_{\alpha}$ can be formalized as: $s_{\alpha} = \alpha s + ( 1 - \alpha ) \hat{s}$.  
In practice, $\hat{s}$ is chosen from \textit{Places Dataset} \cite{places} and $\alpha = 0.5$.
Samples from the above two data augmentations are shown in Fig.\ref{fig_da}.

Since random convolution makes changes to the color of the observation, it allows the agent to adapt to backgrounds with different color transformations and performs better in the \textit{color hard} environment of DMC-GB, while random overlay allows the agent to adapt to backgrounds with different image transformations and thus performs better in \textit{video easy} and \textit{video hard}.

\clearpage
\section{Additional Description and Results of DMC-GB \label{describe_dmcgb}}

\begin{figure}[hbt]
  \centering
  \includegraphics[width=.9\linewidth]{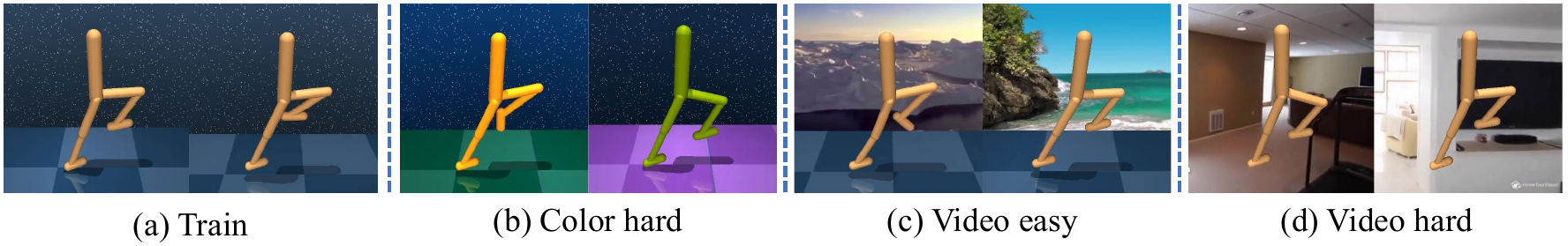}
  \caption{Training and test environments in DMC-GB.}
  \label{fig_dmcgb_env}
\end{figure}

In order to evaluate the agent's generalization abilities, we conducted tests in three environments with different distractions: 
1) \textit{Color hard}: Pixel colors in the image observations are modified. 
2) \textit{Video easy}: Simple videos are added to the background of the image observations, preserving some pixels of the original background. 
3) \textit{Video hard}: Complex videos are added to the background of the image observations, completely covering the entire background. 
The training and testing environments are illustrated in Fig.\ref{fig_dmcgb_env}.

\begin{figure*}[!hbt]
\centering
  \includegraphics[width=\linewidth]{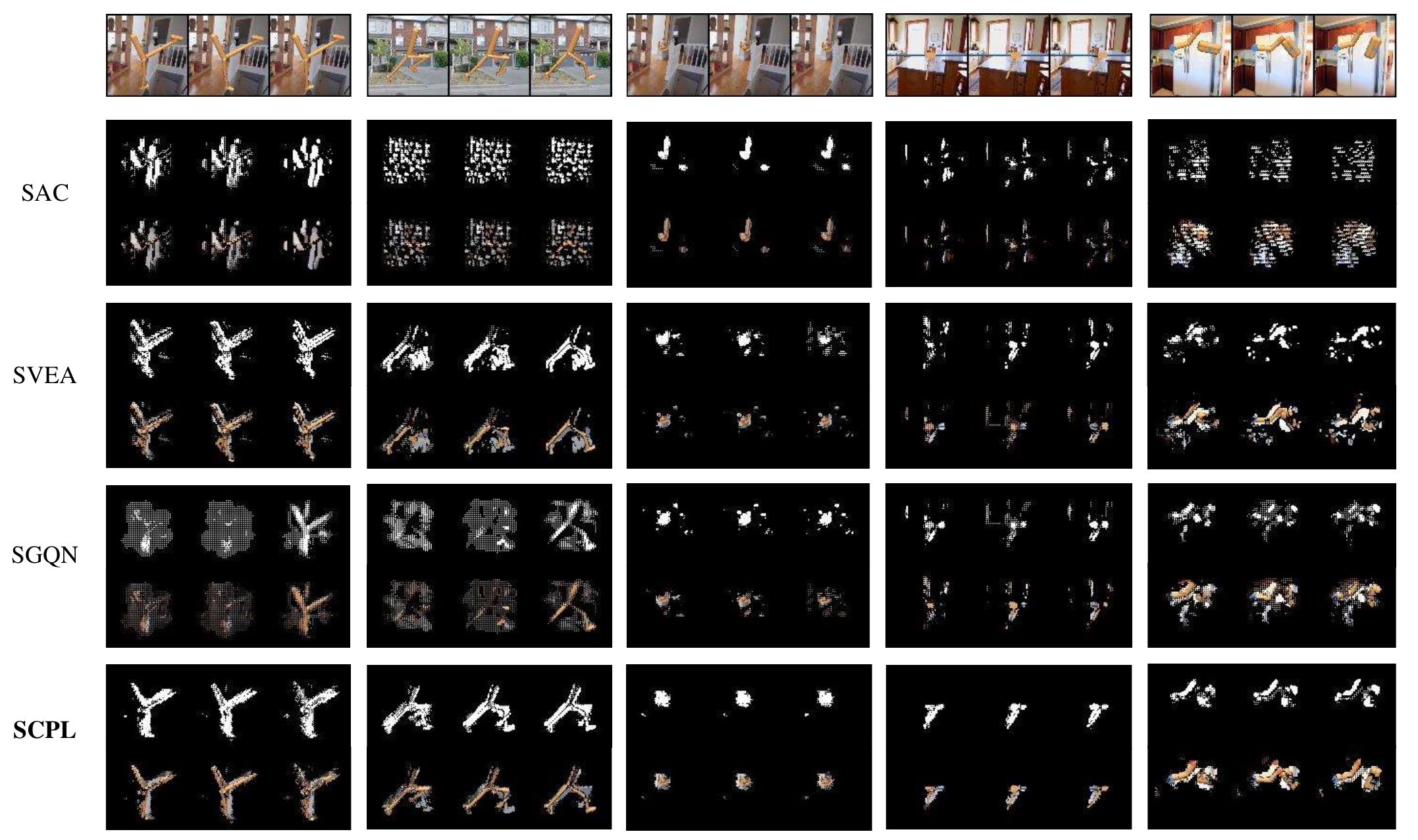}
  \caption{Saliency attribute maps and saliency attribute masked maps in the DMC-GB \textit{video hard} environments. From left to right are the tasks of \textit{walk stand}, \textit{walker walk}, \textit{cartpole swingup}, \textit{ball in cup}, \textit{finger spin}, respectively. The saliency attribute maps (top) and saliency masked maps (bottom) of the Q-network show that
  SCPL can accurately focus on task-relevant regions and discard perturbed pixels in perturbed observations.}
  \label{fig_map_dmc}
\end{figure*}

We also report a comparison of the saliency attribution maps and saliency attribute masked observations for SAC, SVEA, SGQN, and SCPL on all tasks in \textit{video hard} environments in Fig.\ref{fig_map_dmc}. 
The top row in each method represents the saliency attribute maps, while the bottom row displays the saliency attribute masked maps. 
By examining the saliency attribute maps, we can identify the regions that are captured by the agent. 
The saliency attribute masked maps indicate whether the regions focused on by the agent align with the task-relevant pixels.
Among them, SAC struggles to focus on task-relevant regions; SVEA can attend to task-relevant regions but may include more distracting pixels; and SGQN can capture crucial areas in the last frame but cannot consistently perform well across all stacked perturbed observations. 
SCPL precisely attends to task-relevant regions in all stacked perturbed observations.

\begin{figure*}[!hbt]
  \includegraphics[width=\linewidth]{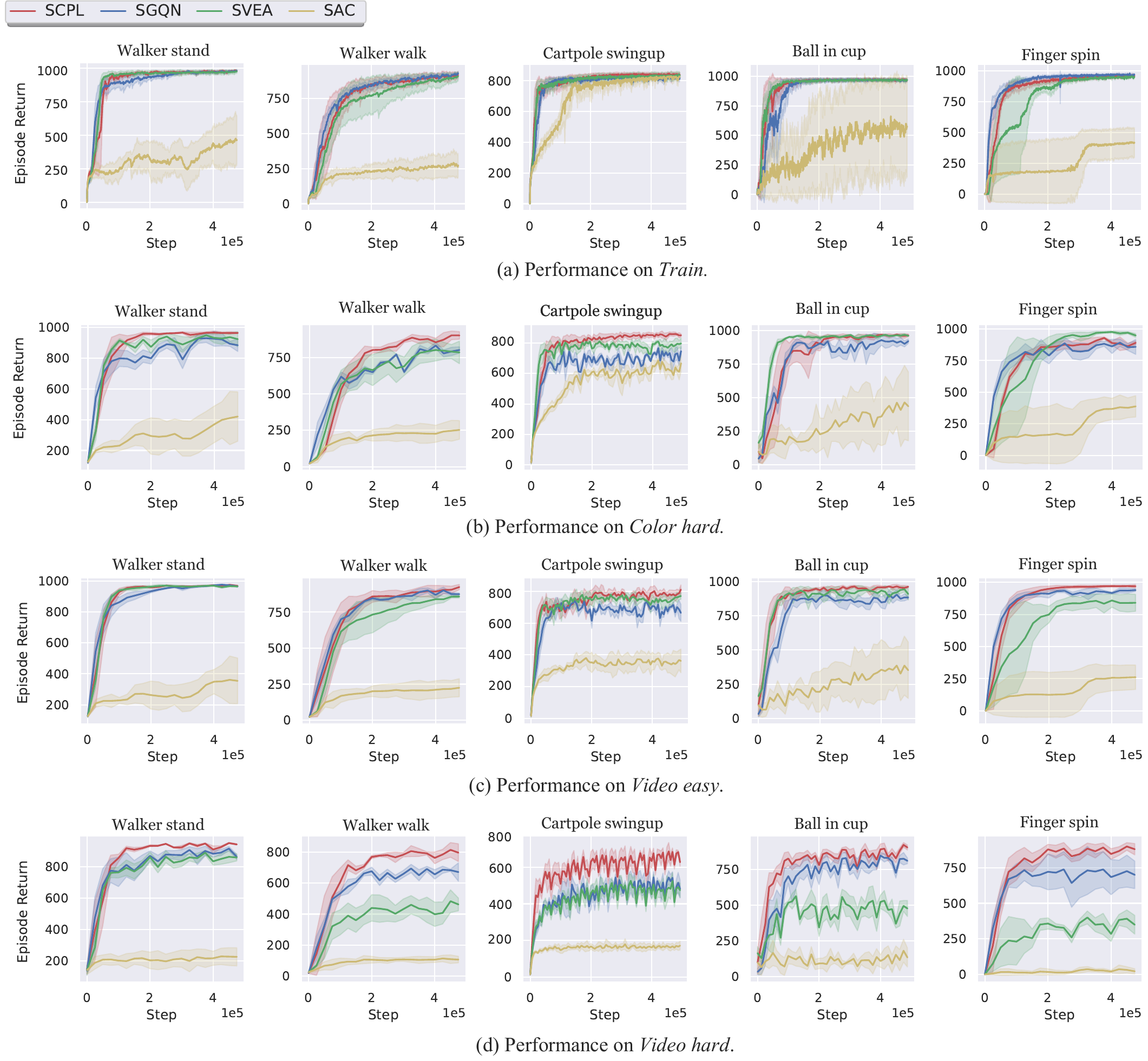}
  \caption{Training and testing curves of SAC, SVEA, SGQN and SCPL in DMC-GB environments.}
  \label{fig_alldmcresult}
\end{figure*}

Fig.\ref{fig_alldmcresult} shows the training curves and testing curves in DMC-GB. 
Data augmentation-based methods often sacrifice data efficiency to enhance generalization, but SCPL achieves similar data efficiency to SOTA methods in the training environment. Across three testing environments, SCPL attains improved generalization for most tasks, particularly in \textit{video hard}, where it outperforms other algorithms by a significant margin across all tasks.

\clearpage
\section{Additional Results of Ablation Study \label{add_ablaction}}
\begin{figure*}[h]
  \includegraphics[width=\linewidth]{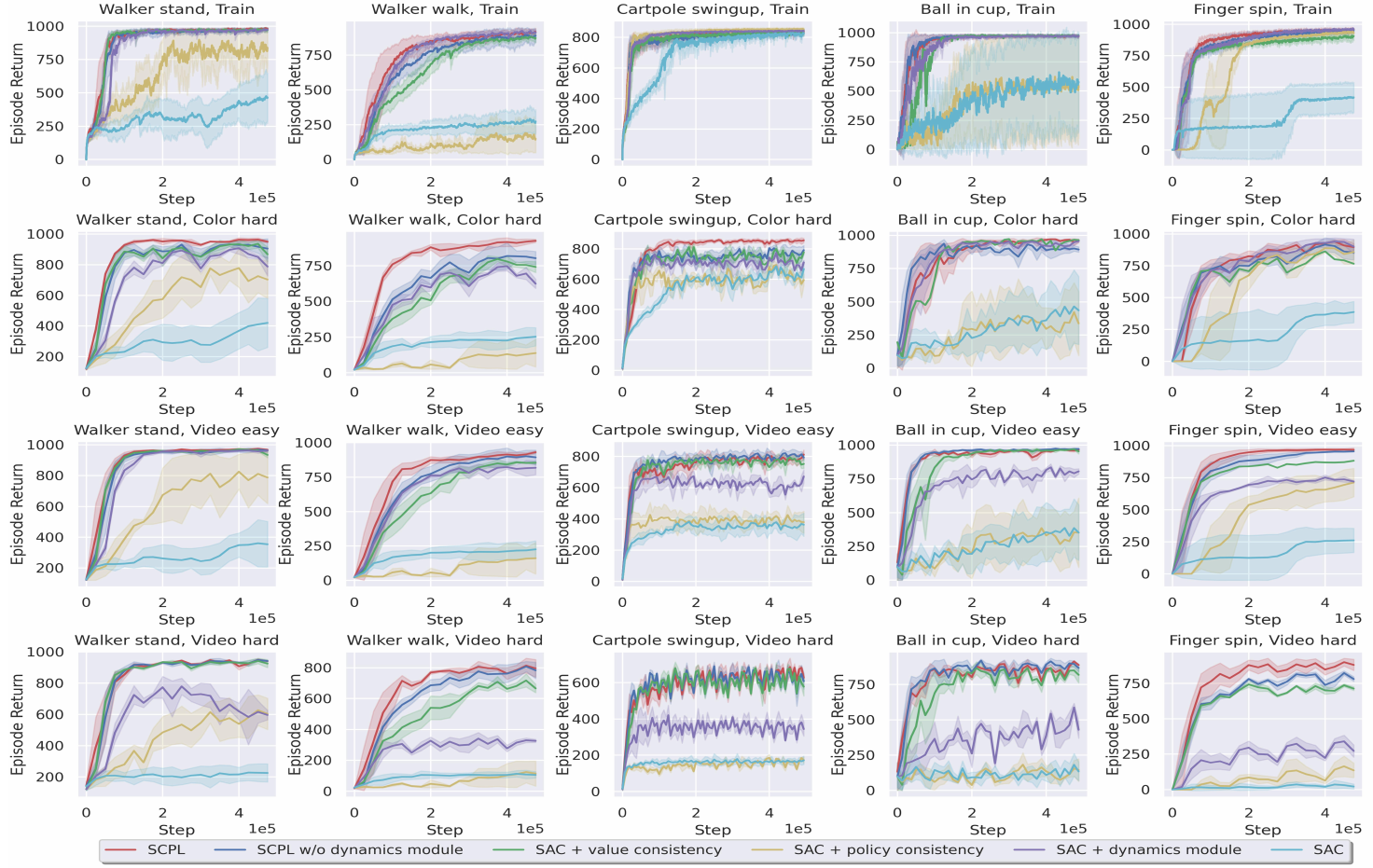}
  \caption{Performance of Ablation Study.}
  \label{fig_alldmcablation}
\end{figure*}

The Fig.\ref{fig_alldmcablation} illustrates the curves of ablation experiments conducted in both training and testing environments. The results of the complete ablation study are presented in Table \ref{table_ablaton_all}. We compared SAC, SAC + value consistency module, SAC + policy consistency module, SAC + dynamics module, SCPL w/o dynamics module, and SCPL. Among these, the SAC + value consistency module, the SAC + policy consistency module, and the SAC + dynamics module are variants that integrate three modules into the SAC baseline. The results depicted in the table demonstrate their improvements in comparison to the SAC algorithm.
SCPL without the dynamics module adds the value consistency module and policy consistency module to the SAC baseline. Their combination in separate modules further enhances the generalization performance. Compared to SCPL without the dynamics module, SCPL introduces the dynamic module, thereby enhancing generalization through the learning of efficient and robust representations.

With these data and curves, we can further illustrate the contribution of our three modules.
The value consistency module achieves the greatest improvement in algorithm generalization by precisely directing the value function's attention to task-relevant regions within the observations. Policy consistency enhances generalization by encouraging the policy network to generate actions consistent with the original observations for various perturbed observations. 
The dynamics module provides efficient and robust representations for the value function and policy network by predicting the representations and rewards of the next state for both original and augmented observations, further enhancing generalization performance.

\begin{table*}[h]
\centering
\begin{tabular}{cccccccc}
\toprule
\thead{Setting} & Task & SAC  & \thead{SAC + \\ policy consistency} & \thead{SAC + \\ dynamics module} & \thead{SAC + \\ value consistency} & \thead{SCPL w/o \\ dynamics module } & \textbf{SCPL} \\
\midrule
\multirow{6}{*}{\thead{Color \\ hard} } &
Walker stand &
$423 \pm 155$ &
$733 \pm 88$ &
$841 \pm 6$ &
$838 \pm 57$ &
$952 \pm 20$ &
$\textbf{957} \pm \textbf{11}$  \\
 &
Walker walk &
$255 \pm 61$ &
$165 \pm 86$ &
$703 \pm 56$ &
$751 \pm 28$ &
$821 \pm 38$ &
$\textbf{845} \pm \textbf{17}$  \\
 &
Cartpole &
$615 \pm 29$ &
$644 \pm 24$ &
$733 \pm 3$ &
$793 \pm 2$ &
$795 \pm 4$ &
$\textbf{815} \pm \textbf{6}$ \\
 &
Ball in cup &
$391 \pm 245$ &
$398 \pm 152$ &
$\textbf{960} \pm \textbf{2}$ &
$923 \pm 40$ &
$897 \pm 23$ &
$955 \pm 12$  \\
 &
Finger spin &
$373 \pm 70$ &
$865 \pm 102$ &
$886 \pm 70$ &
$734 \pm 22$ &
$883 \pm 38$ &
$\textbf{901} \pm \textbf{38}$ \\ \cline{2-8}
 &
Average &
$411$ &
$561(+\textbf{36}\%)$ &
$825(+\textbf{101}\%)$ &
$808(+\textbf{97}\%)$ &
$870(+\textbf{112}\%)$ &
$\textbf{893}(+\textbf{117}\%)$ \\

\midrule
\multirow{6}{*}{\thead{Video \\ easy} } &
Walker stand &
$351 \pm 145$ &
$835 \pm 126$ &
$956 \pm 16$ &
$921 \pm 22$ &
$968 \pm 5$ & 
$\textbf{969} \pm \textbf{8}$ \\
 &
Walker walk &
$228 \pm 65$ &
$197 \pm 96$ &
$816 \pm 70$ &
$851 \pm 7$ &
$903 \pm 45$ & 
$\textbf{941} \pm \textbf{9}$  \\
 &
Cartpole &
$359 \pm 80$ &
$371 \pm 68$ &
$691 \pm 37$ &
$754 \pm 33$ &
$801 \pm 27$ & 
$\textbf{814} \pm \textbf{21}$ \\
 &
Ball in cup &
$338 \pm 201$ &
$449 \pm 203$ &
$811 \pm 15$ &
$947 \pm 27$ &
$\textbf{974} \pm \textbf{3}$ & 
$963 \pm 10$ \\
 &
Finger spin &
$260 \pm 98$ &
$720 \pm 108$ &
$712 \pm 4$ &
$899 \pm 26$ &
$949 \pm 7$ & 
$\textbf{963} \pm \textbf{8}$  \\ \cline{2-8}
 &
Average &
$300$ &
$514(+\textbf{71}\%)$ &
$797(+\textbf{166}\%)$ &
$874(+\textbf{191}\%)$ &
$919(+\textbf{206}\%)$ &
$\textbf{930}(+\textbf{210}\%)$  \\
\midrule
\multirow{6}{*}{\thead{Video \\ hard} } &
Walker stand &
$225 \pm 58$ &
$647 \pm 83$ &
$630 \pm 26$ &
$918 \pm 29$ &
$949 \pm 9$ &
$\textbf{953} \pm \textbf{15}$ \\
 &
Walker walk &
$104 \pm 18$ &
$132 \pm 65$ &
$336 \pm 14$ &
$673 \pm 9$ &
$812 \pm 20$ &
$\textbf{818} \pm \textbf{32}$ \\
 &
Cartpole &
$174 \pm 24$ &
$194 \pm 5$ &
$351 \pm 15$ &
$567 \pm 64$ &
$624 \pm 52$ &
$\textbf{675} \pm \textbf{3}$ \\
 &
Ball in cup &
$196 \pm 82$ &
$237 \pm 9$ &
$405 \pm 29$ &
$805 \pm 67$ &
$909 \pm 12$ &
$\textbf{924} \pm \textbf{7}$  \\
 &
Finger spin &
$26 \pm 21$ &
$168 \pm 87$ &
$292 \pm 6$ &
$703 \pm 15$ &
$802 \pm 6$ &
$\textbf{897} \pm \textbf{22}$ \\ \cline{2-8}
 &
Average &
$145$ &
$275(+\textbf{90}\%)$ &
$403(+\textbf{178}\%)$ &
$733(+\textbf{405}\%)$ &
$819(+\textbf{465}\%)$ &
$\textbf{853}(+\textbf{488}\%)$ \\
\bottomrule
\end{tabular}
\caption{Ablation study. We compare the test performance of SAC with the different modules included, illustrating the effectiveness of each module for performance improvement. Percentages indicate variations compared to vanilla SAC.}
\label{table_ablaton_all}
\end{table*}

\clearpage
\section{Additional Description and Results of Robotic Environment \label{add_robot}}
\begin{figure}[!hbt]
\centering
  \includegraphics[width=.55\linewidth]{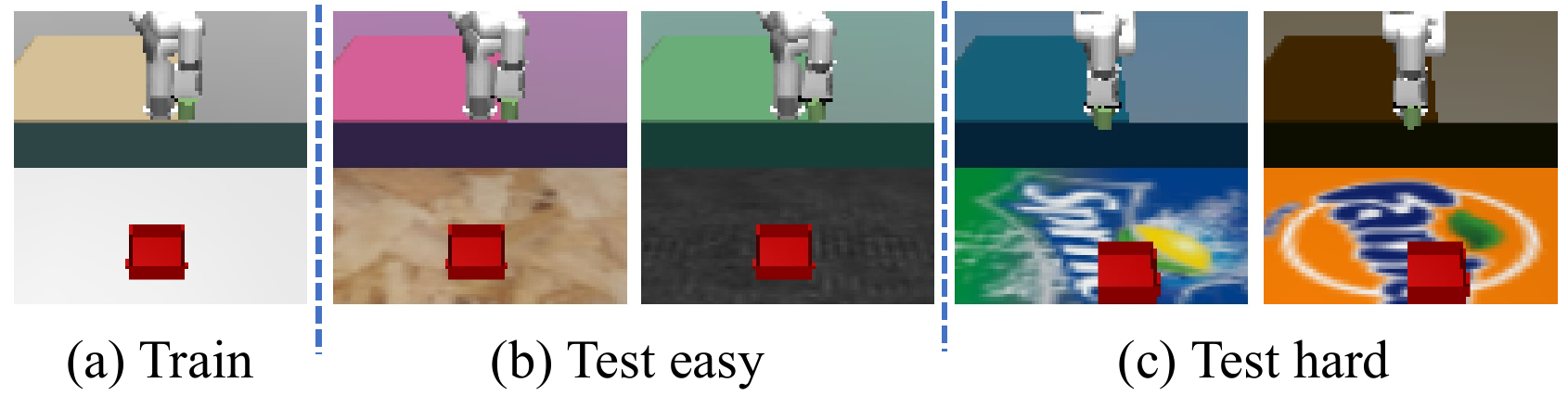}
  \caption{Vision-based Robotic Manipulation Environments. This figure shows examples of training and testing observation for the \textit{Peg in Box} task.}
  \label{fig_robot_env}
\end{figure}

We carried out experiments on Robotic Manipulation with three tasks:
(a)\textit{Reach}: a task where the robot has to reach a randomly positioned mark on the table with the robot's end-effector. 
(b)\textit{Push}: a task where the robot has to push a box to a target position indicated by a mark on the table. 
(c)\textit{Peg in Box}: a task where the robot has to place a peg attached to the robot's end-effector with a string into a box. 
Fig.\ref{fig_robot_env} shows the training and test environments of our Robotic environments.
During training, the agent is trained with a default desktop and background.
In the easy mode of test environments, the agent is evaluated under different colors and desktop textures.
In the hard mode of test environments, the agent is tested under complex desktop backgrounds with intricate images.

\begin{figure*}[!hbt]
\centering
  \includegraphics[width=.85\linewidth]{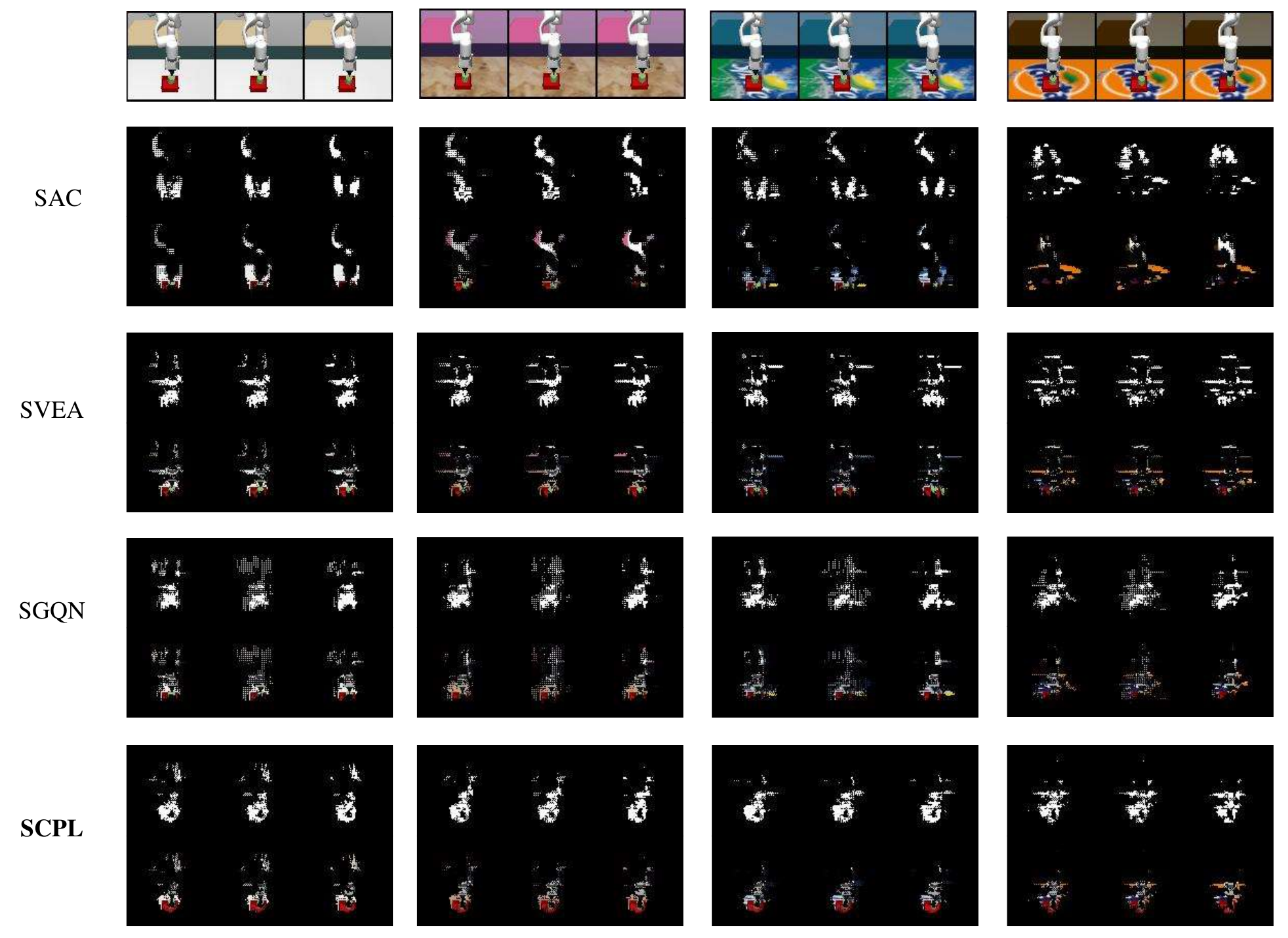}
  \caption{Saliency attribute maps and saliency attribute masked maps in the Robotic Environment, \textit{peg in box} tasks. From left to right are the  training environments, easy-mode test environments, and two hard-mode environments. The saliency attribute
maps (top) and saliency masked maps (bottom) show that SCPL can accurately focus on the ball and box that need to be picked up in the presence of strong interference. }
  \label{fig_map_robot}
\end{figure*}

The saliency masked maps of SAC, SVEA, SGQN, and SCPL in different test environments are shown in Fig.\ref{fig_map_robot}. The results show that SCPL can accurately focus on the ball and box that need to be picked up in the presence of strong perturbations, while SVEA and SGQN have poor capture.

\clearpage
\section{Additional description and results of CARLA \label{add_carla}}
\begin{figure}[!hbt]
\centering
  \includegraphics[width=0.5\linewidth]{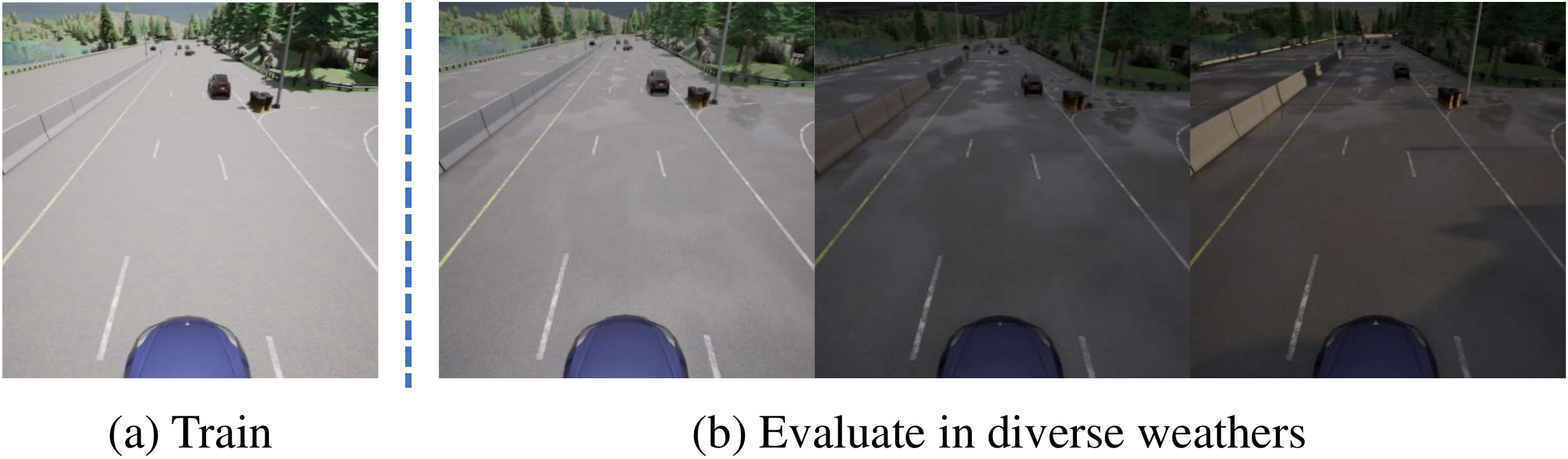}
  \caption{CARLA environments for generalization. Agents are trained in the weather of \textit{clear noon} and evaluated in diverse weather conditions.}
  \label{fig_carla_env}
  \vspace{-.2cm}
\end{figure}

In CARLA experiments, the vehicle is trained in the  weather of \textit{clean noon}, and  tested  in  \textit{wet cloudy noon}, \textit{hard rain noon}, \textit{wet sunset}, \textit{soft rain sunset}, \textit{mid rain sunset}, respectively. The training and test environments are shown in Fig.\ref{fig_carla_env}.

In the reward function, to avoid the ego vehicle getting stuck, we use a low speed penalty term, i.e., $r_v=-100$ (if $\mathbf{v}_{agent} < 0.02 m/s$ and $time>10s$) or $r_v=0$ (otherwise). In order to improve safety, a collision penalty term is set to $r_c=-100$ (if collision happens) or $r_c=0$ (otherwise). The episode is terminated early if $r_v=-100$ or $r_c=-100$ to improve the efficiency of RL algorithms. The remaining part of the reward is $r_{t}=\mathbf{v}_{agent}^{\top} \hat{\mathbf{u}}_{highway} \cdot \Delta t-\mid steer \mid$, where $\mathbf{v}_{agent}$ is the velocity vector of the ego vehicle, and the dot product with the unit vector of the highway $\mathbf{u}_{highway}$ encourages progression along the highway as fast as possible. $\Delta t = 0.05$ discretizes the simulation time and $steer \in [-1,1]$ denotes the normalized steering wheel angle. Finally, the total reward is $R = r_v + r_c + r_t$.

\begin{figure*}[!hbt]
\centering
  \includegraphics[width=.8\linewidth]{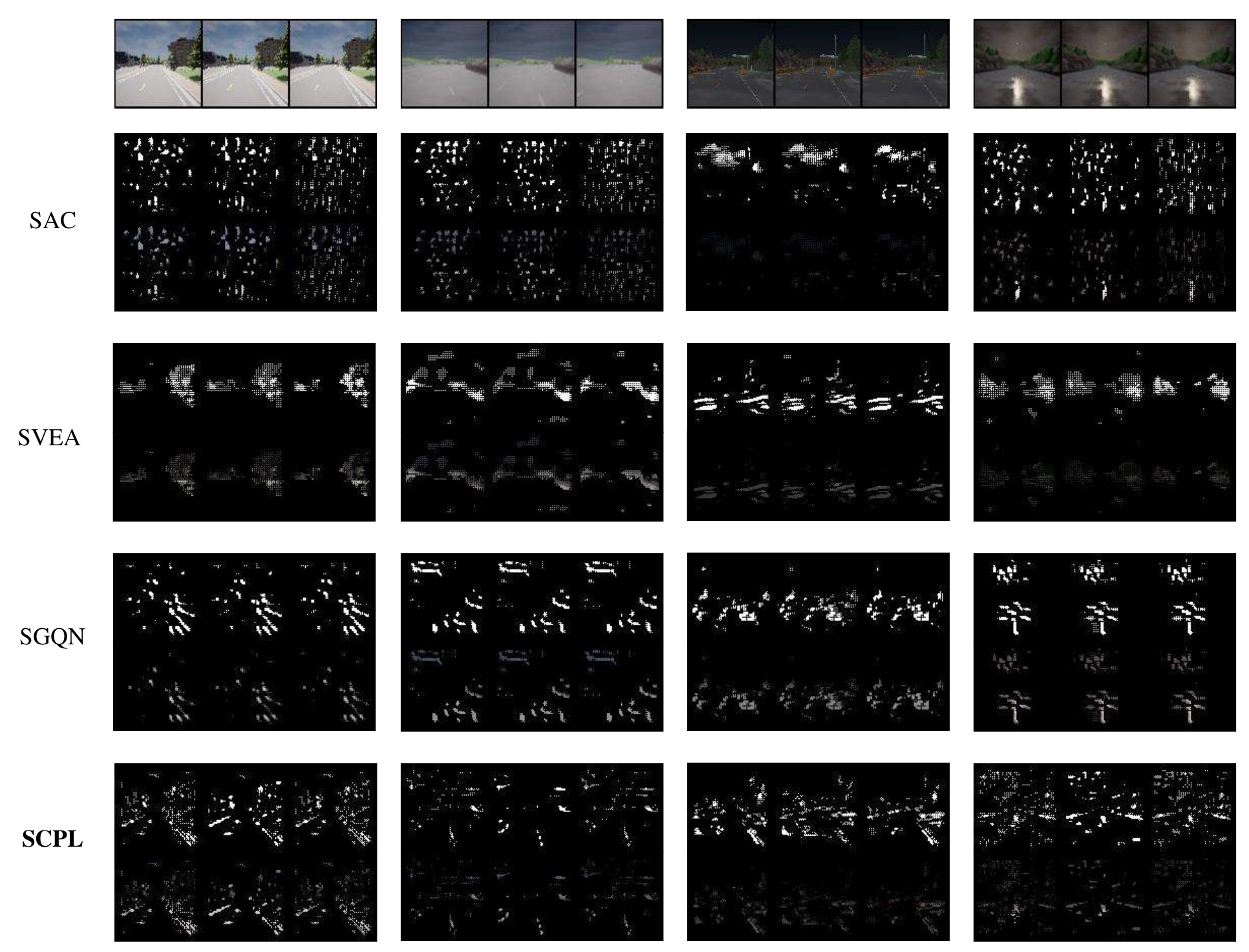}
  \caption{Saliency attribute maps and saliency attribute masked maps of SAC, SVEA, SGQN, and SCPL in CARLA. The weather of environments from left to right are \textit{clean noon}, \textit{wet sunset}, \textit{soft rain sunset}, and \textit{mid rain sunset}. SCPL can recognize lane line information in most weather conditions.}
  \label{fig_map_carla}
\end{figure*}

The saliency masked maps of SAC, SVEA, SGQN, and SCPL in Carla environments are shown in Fig.\ref{fig_map_carla}. It represents that SCPL can recognize lane line information in the presence of interference, while other algorithms fail to do so.

\clearpage
\section{Hyperparameter}
Hyperparameters of all environments in our experiences are shown in Table \ref{table_hyper}.

\begin{table*}[hbt]
\centering
\begin{tabular}{c|ccc}
\toprule
Hyperparameter  & DMC & Robot & CARLA \\
\midrule
Input dimension &
$9 \times 84 \times 84$ &
$9 \times 84 \times 84$ &
$9 \times 84 \times 84$  \\
Stacked frames &
$3$ &
$3$ &
$3$  \\
Discount factor $\gamma$ &
$0.99$ &
$0.99$ &
$0.99$  \\
Episode length &
$1000$ &
$50$ &
$1000$  \\
Training steps &
$500K$ &
\thead{$200K$(push) \\ $100K$(otherwise)} &
$200K$  \\
Replay buffer size &
$100K$ &
$100K$ &
$100K$  \\
Batch size &
$128$ &
$128$ &
$128$  \\
Action repeat &
\thead{$8$(cartpole) $4$(ball\_in\_cup) \\ $2$(otherwise)} &
$1$ &
$2$  \\
Actor learning rate &
$1e-3$  &
$1e-3$ &
$1e-3$  \\
Critic learning rate &
$1e-3$ &
$1e-3$ &
$1e-3$  \\
$\log \alpha$ learning rate &
$1e-4$ &
$1e-4$ &
$1e-4$  \\
Target network update frequency &
$2$ &
$2$ &
$2$  \\
Encoder conv layers &
$11$ &
$11$ &
$4$  \\
Encoder feature dim &
$100$ &
$100$ &
$100$  \\
Random cropping padding &
$4$ &
$4$ &
$4$  \\
Optimizer($\theta$) &
$Adam$ &
$Adam$ &
$Adam$  \\
Quantile value $\rho$ &
$0.9$(walker) $0.95$(otherwise) &
$0.95$ &
$0.9$  \\
KL loss coefficient $\beta$ &
$1$ &
$1$ &
$1$  \\
 value consistency coefficient $\lambda$ &
$0.5$ &
$0.5$ &
$0.5$  \\
\bottomrule
\end{tabular}
\caption{SCPL hyperparameters for all environments.}
\label{table_hyper}
\end{table*}

\clearpage
\section{Impact of $\rho$}
\begin{figure*}[!hbt]
\centering
  \includegraphics[width=.9\linewidth]{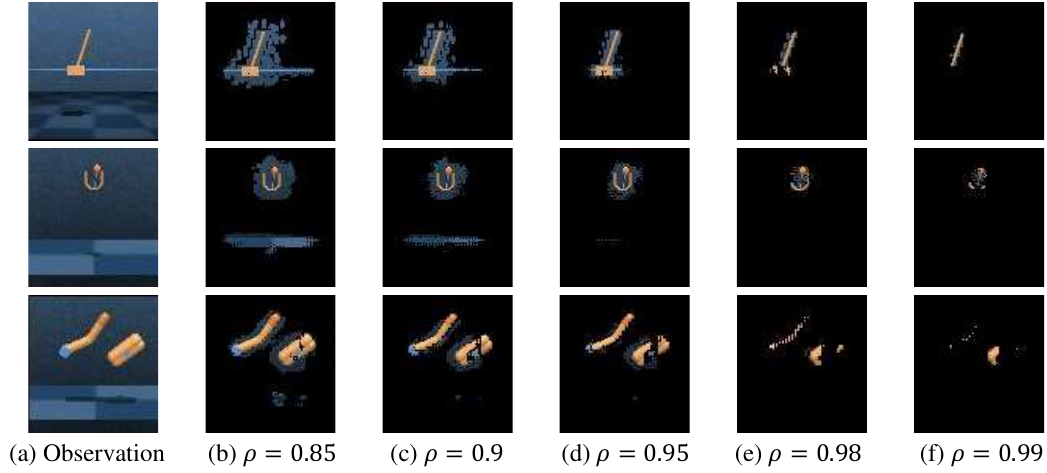}
  \caption{Example of $M_{\rho}(Q_{\theta},s,a)$ with different values of $\rho$ on \textit{cartpole swing up} (top), \textit{ball in cup} (middle), and \textit{finger spin} (bottom). }
  \label{compare_rho1}
\end{figure*}

In our experiments, we utilize a rapid visual search approach to determine the hyperparameter value, $\rho$. The saliency maps obtained with different $\rho$ values are depicted in Fig.\ref{compare_rho1}. In the DMC-GB experiments, due to the larger task-relevant regions in \textit{walker stand} and \textit{walker walk}, we employ a $\rho$ value of 0.9. For other tasks, we used a $\rho$ value of 0.95. To evaluate SCPL's sensitivity to $\rho$, we compared the performance of \textit{cartpole swing up}, \textit{ball in cup}, and \textit{finger spin} tasks with $\rho$ values of 0.9 and 0.95. From Table \ref{table_rho}, it can be observed that the $\rho$ value of 0.95 slightly outperforms 0.9 in the three tasks. Due to the varying shape of task-relevant regions in different tasks, it is necessary to fine-tune the parameter $\rho$ for different tasks.

\begin{table*}[hbt]
\centering
\resizebox{0.4\columnwidth}{!}{
\begin{tabular}{llll}
\toprule
\thead{Setting} & Task & $\rho=0.9$ & $\rho=0.95$ \\
\midrule
\multirow{3}{*}{\thead{Color \\ hard}} & 
Cartpole  &
$815 \pm 6$ &  
$816 \pm 35$ \\
 &
Ball in cup &
$955 \pm 12$ &
$952 \pm 18$\\
 &
 Finger spin &
 $901 \pm 38$&
 $902 \pm 33$\\ 
\midrule
\multirow{3}{*}{\thead{Video \\ easy}} &
Cartpole  &
$814 \pm 21$ &
$831 \pm 28$ \\
 &
Ball in cup &
$963 \pm 10$ &
$963 \pm 14$ \\
 &
Finger spin &
$963 \pm 8$ &
$974 \pm 7$\\
\midrule
\multirow{3}{*}{\thead{Video \\ hard}} &
Cartpole  &  
$675 \pm 3$ &  
$685 \pm 5$ \\
 &
Ball in cup &  
$921 \pm 30$ &  
$924 \pm 7$ \\
 &
Finger spin &
$873 \pm 22$ & 
$897 \pm29$ \\
\bottomrule
\end{tabular}
}
\caption{ Impact of $\rho$ for \textit{cartpole swing up}, \textit{ball in cup}, and \textit{finger spin}.}
\label{table_rho}
\end{table*}

\clearpage
\section{Discussion on direct representation learning}
Doing representation learning directly (such as SODA) is a straightforward method for the generalization of VRL.
SODA\cite{overlay}, which ensures that the features are similar for data with or without augmentation. While this approach might seem viable, it carries the risk of overfitting. As shown in Fig. 8, the SAC agent tend to focus on regions containing task-irrelevant pixels in the original observations. Aligning representations directly could lead to similar attention regions in augmented observations, which is not desirable. Although this might have a minimal impact in training environments where agents overfit to the original observations, it could significantly hinder performance in more complex or perturbed settings.

To evaluate the generalization ability of SCPL, we compare its generalization performance with SODA in DMC-GB, as presented in Table \ref{table_comparison}. SCPL consistently outperforms SODA in terms of generalization across both environments. These experimental results highlight the advantages of the components incorporated in SCPL.

\begin{table*}[hbt]
\centering
\resizebox{0.7\columnwidth}{!}{
\begin{tabular}{llllllll}
\toprule
Setting & Method & Walker Stand & Walker Walk & Cartpole & Ball & Finger & Average \\
\midrule
\multirow{2}{*}{Video Easy} &
SODA & 965 & 771 & 742 & 939 & 783 & 836 \\
& SCPL & 968 & 941 & 814 & 963 & 963 & 930 \\
\midrule
\multirow{2}{*}{Video Hard} &
SODA & 736 & 312 & 339 & 381 & 309 & 430 \\
& SCPL & 953 & 818 & 675 & 924 & 897 & 853 \\
\bottomrule
\end{tabular}
}
\caption{Comparison of SODA and SCPL for various tasks under different settings.}
\label{table_comparison}
\end{table*}

\end{document}